\documentclass[sigconf]{acmart}

\renewcommand\footnotetextcopyrightpermission[1]{}
\usepackage{algorithm}
\usepackage{algpseudocode}
\usepackage{amsmath}
\usepackage{amssymb}  
\usepackage{multirow} 

\usepackage{graphicx}
\usepackage{booktabs}
\usepackage{caption}
\usepackage{subcaption}
\usepackage{hyperref}
\usepackage{float}

\usepackage{adjustbox}
\usepackage{url}

\acmConference[Preprint]{arXiv}{arXiv}{2026} 
\acmYear{}
\copyrightyear{}

\begin{document}

\title[engGNN]{engGNN: A Dual-Graph Neural Network for Omics-Based Disease Classification and Feature Selection}

\author{Tiantian Yang}
\affiliation{
  \institution{Department of Mathematics and Statistical Science, University of Idaho}
  \city{Moscow}
  \state{Idaho}
  \country{USA}
}
\affiliation{
  \institution{Department of Biostatistics, Boston University}
  \city{Boston}
  \state{Massachusetts}
  \country{USA}
}
\email{tyang@uidaho.edu}

\author{Yuxuan Wang}
\affiliation{
  \institution{Department of Biostatistics, Boston University}
  \city{Boston}
  \state{Massachusetts} 
  \country{USA}
}      

\author{Zhenwei Zhou}
\affiliation{
  \institution{Department of Biostatistics, Boston University}
  \city{Boston}
  \state{Massachusetts} 
  \country{USA}
}      

\author{Ching-Ti Liu}
\affiliation{
  \institution{Department of Biostatistics, Boston University}
  \city{Boston}
  \state{Massachusetts} 
  \country{USA}
} 
\email{ctliu@bu.edu}

\renewcommand{\shortauthors}{Yang et al.}

\begin{abstract}
Omics data, such as transcriptomics, proteomics, and metabolomics, provide critical insights into disease mechanisms and clinical outcomes. However, their high dimensionality, small sample sizes, and intricate biological networks pose major challenges for reliable prediction and meaningful interpretation. 
Graph Neural Networks (GNNs) offer a promising way to integrate prior knowledge by encoding feature relationships as graphs. Yet, existing methods typically rely solely on either an externally curated feature graph or a data-driven generated one, which limits their ability to capture complementary information. To address this, we propose the \textbf{e}xternal a\textbf{n}d \textbf{g}enerated \textbf{G}raph \textbf{N}eural \textbf{N}etwork (\textbf{engGNN}), a dual-graph framework that jointly leverages both external known biological networks and data-driven generated graphs. Specifically, engGNN constructs a biologically informed undirected feature graph from established network databases and complements it with a directed feature graph derived from tree-ensemble models. This dual-graph design produces more comprehensive embeddings, thereby improving predictive performance and interpretability. Through extensive simulations and real-world applications to gene expression data, engGNN consistently outperforms state-of-the-art baselines. Beyond classification, engGNN provides interpretable feature importance scores that facilitate biologically meaningful discoveries, such as pathway enrichment analysis. Taken together, these results highlight engGNN as a robust, flexible, and interpretable framework for disease classification and biomarker discovery in high-dimensional omics contexts.
\end{abstract}

\maketitle

\section{Introduction \label{section intro}}

The rapid growth of omics data, such as genomics, transcriptomics, and proteomics, provides unique opportunities to uncover the regulatory mechanisms of diseases and holds great promise for predicting clinical outcomes \cite{hood2013human, hasin2017multi}. For example, gene expression profiles and gene networks have been utilized to amplify signals from genomic data, elucidate disease mechanisms, facilitate diagnostic efforts, detect risk variants, and identify biomarkers \cite{van2002gene, subramanian2005gene, barabasi2011network}. These data capture complex relationships among biological entities and processes, yet their high dimensionality and typically limited sample sizes present substantial challenges for building robust predictive models. In addition, omics data introduce complexities, such as sparsity and uncertainty within biological networks. Unknown interactions between features, such as gene expression profiles, further complicate classification tasks. Importantly, interpretability remains a critical and unresolved issue in the development of reliable models \cite{berger2013computational, tarazona2021undisclosed}. 

While conventional machine learning (ML) techniques excel in classification and prediction tasks, they often struggle to process natural data in raw form \cite{lecun2015deep}. 
Deep learning (DL), as a representation learning framework, has shown great promise in automatically extracting meaningful patterns from raw data, with notable success across diverse domains ranging from image and speech recognition to natural language processing \cite{lecun2015deep, krizhevsky2017imagenet, hinton2012deep, collobert2011natural}. 
In biomedical research, DL has been applied to tasks such as predicting drug activity \cite{ma2015deep}, assessing the effects of mutation \cite{leung2014deep}, and diagnosing diseases \cite{xiong2015human}. However, the `large $p$, small $n$' setting, where the number of features ($p$) far exceeds the number of samples ($n$) \cite{ebbinghaus2005less}, limits the effectiveness of DL models, which generally rely on large-scale training data. 

Incorporating functional relationships among features through graph structures offers a natural and principled way to represent dependencies and introduce informative priors \cite{bruna2014spectral, henaff2015deep}. Graph representation learning (GRL) leverages these structures to learn embeddings of node-level attributes and their interactions \cite{hamilton2017representation}. Graph neural networks (GNNs) have recently emerged as powerful tools for solving GRL problems \cite{zhang2020deep, wu2020comprehensive, zhou2020graph}, where feature information is propagated and aggregated along graph edges to improve predictive performance \cite{ying2019gnn}. Depending on the model architectures and training strategies, state-of-the-art GNN methods can be categorized into various types, including graph recurrent neural networks, graph convolutional networks, graph autoencoders, graph reinforcement learning, and graph adversarial methods \cite{zhang2020deep}. 
Previous studies have shown that incorporating graph-based priors improves classification in biomedical studies \cite{zhang2021graph, paul2024systematic, johnson2024graph}. 

However, most existing GNN-based models face a persistent limitation that they rely on a single source of graph information: either an external biological network (e.g., protein-protein interaction networks) or a graph constructed from feature data (e.g., based on similarity or supervised learning). 
For example, the DMGCN model incorporates multiple gene networks as external priors to reduce bias from a single graph \cite{yang2021graph}, but still depends entirely on external biological networks. Conversely, forgeNet builds a task-specific graph using tree ensembles \cite{kong2020forgenet} while MLA-GNN constructs co-expression graphs via weighted correlation network analysis from transcriptomic data \cite{xing2022multi}. Other models, such as \cite{ji2024inferring}, use graph convolutional networks to infer regulatory networks directly from expression data. 
Each of these graph-building strategies provides only a partial view: external graphs from network databases may be incomplete or misannotated, similarity-based graphs may overfit the expression data, and model-generated graphs may encode task-specific noise. Moreover, most existing approaches utilize only one type of graph, either directed or undirected, thereby limiting the scope of structural information that can be learned. 
Another major challenge in applying GNNs to biomedical data is interpretability. Neural networks are often criticized as `black boxes' that obscure the contribution of individual features, and this issue is further amplified in GNNs due to the complex interdependencies among graph nodes and edges \cite{zhang2020deep, wu2020comprehensive, ying2019gnn, olden2002illuminating, olden2004accurate}. This lack of transparency limits their practical utility in biomedical research, where interpretability is essential to uncover disease mechanisms and identify clinically relevant biomarkers.

To address these challenges, we propose the \textbf{e}xternal a\textbf{n}d \textbf{g}enerated \textbf{G}raph \textbf{N}eural \textbf{N}etwork (engGNN) model. The engGNN model integrates feature information from both external biological networks and data-driven feature graphs, effectively combining prior knowledge with novel interactions learned directly from data. By leveraging both undirected and directed graph structures, engGNN enhances representational capacity and model performance. Furthermore, to enhance interpretability, engGNN provides feature importance scores that can guide downstream analyses, such as pathway enrichment. 

\section{Materials and Methods}
This section describes the architecture, algorithm, and implementation details of the proposed engGNN framework, along with the procedure used to rank feature importance scores. \\ 
\noindent \textbf{Preliminaries:} A known graph of $p$ features can be denoted as $G = (V, E, A)$, where $V$ is the set of nodes, $E$ is the set of edges connecting the nodes, and $A \in \mathbb{R}^{p\times p}$ is the adjacency matrix describing the connection strength between nodes. If two nodes are neighbors, the corresponding entry in $A$ is set to 1; otherwise, it is set to 0. A modified adjacency matrix with self-loops is obtained as $\tilde{A} = A + I_{p}$, where $I_{p}$ is the identity matrix. 

\subsection{Overview of the engGNN Framework}
The \textbf{e}xternal a\textbf{n}d \textbf{g}enerated \textbf{G}raph \textbf{N}eural \textbf{N}etwork (engGNN) framework combines prior biological knowledge with data-driven feature graphs. Specifically, it utilizes curated network databases to construct an external undirected feature graph and employs XGBoost to generate a directed feature graph. The model consists of four main components: (1) a graph-embedded deep feedforward network (GEDFN) \cite{kong2018graph} model that aggregates information from an external undirected feature graph, (2) another GEDFN model that aggregates information from a generated directed feature graph, (3) a concatenation of the outputs from the two GEDFN models, and (4) a deep feedforward network (DFN) \cite{goodfellow2016deep} model that learns from the concatenated representation and produces the final prediction.

\begin{figure*}[h!]
\centering
    \addtolength{\leftskip} {-2.6cm} 
    \addtolength{\rightskip}{-2.6cm} 
\includegraphics[width=0.9\textwidth]{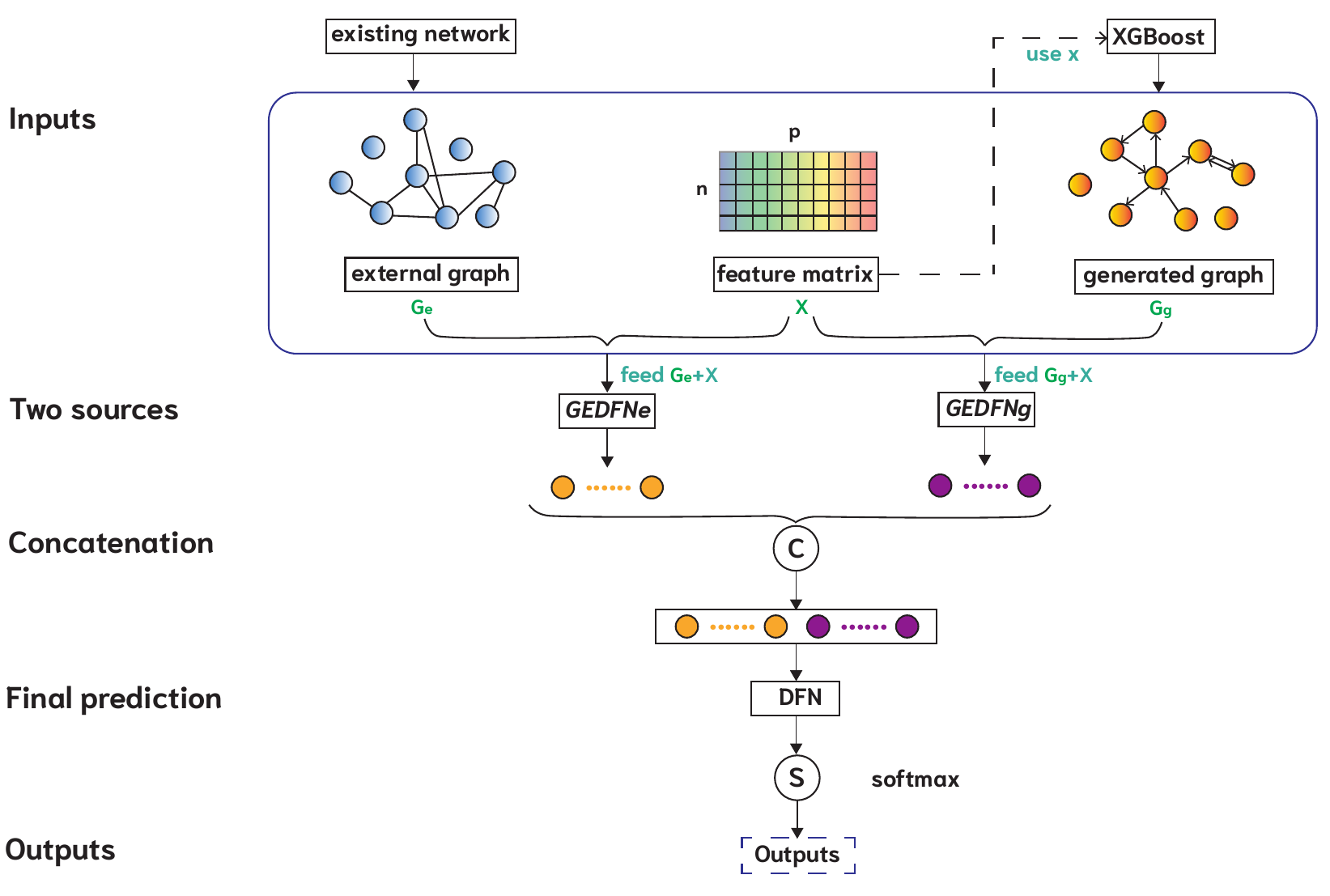}
\caption{Architecture of the proposed \textbf{e}xternal a\textbf{n}d \textbf{g}enerated \textbf{G}raph \textbf{N}eural \textbf{N}etwork (engGNN). The inputs include feature matrix $X$, an external undirected feature graph $G_e$, and a generated directed feature graph $G_{g}$. Step 1: $X$ and $G_{e}$ are processed by a graph-embedded deep feedforward network ($\mathrm{GEDFN}_{\text{e}}$); Step 2: $X$ and  $G_{g}$ are processed by another GEDFN ($\mathrm{GEDFN}_{\text{g}}$); Step 3: the last hidden layers of $\mathrm{GEDFN}_{\text{e}}$ and $\mathrm{GEDFN}_{\text{g}}$ are concatenated; Step 4: the concatenated representation is passed through a deep feedforward network (DFN), followed by a softmax activation to generate the final predicted class probabilities $\hat{Y}$. Overviews of $\mathrm{GEDFN}_{\text{e}}$ and $\mathrm{GEDFN}_{\text{g}}$ are detailed in Figure~\ref{GEDFN_e} and Figure~\ref{GEDFN_g}, respectively.
\label{engGNN}}
\end{figure*}

Algorithm \ref{algo1} summarizes the procedure, and Figure \ref{engGNN} illustrates the overall architecture. The inputs include the feature matrix $X$ (e.g., gene expression profiles), the external undirected feature graph $G_e$, and the generated directed feature graph $G_{g}$. Each graph is processed by a separate GEDFN model (denoted as $\mathrm{GEDFN}_{\text{e}}$ and $\mathrm{GEDFN}_{\text{g}}$), producing embeddings $H_{G_{e}}$ and $H_{G_{g}}$, respectively. These embeddings capture complementary information and are concatenated into a single representation: $H_{C} = [H_{G_{e}} \,\|\, H_{G_{g}}]$. The concatenated representation $H_{C}$ is then passed to a DFN model for final classification.   
\begin{algorithm}[h]
\caption{\textbf{engGNN}}
\label{algo1}
\begin{algorithmic}[1]
\Require Feature matrix $X$, external feature graph $G_{e}$, generated feature graph $G_{g}$ 
\Ensure Predicted binary class labels $\hat{Y}$ 
    \State Compute embedding from external undirected graph: $H_{G_{e}} = \text{GEDFN}(X, G_{e})$
    \State Compute embedding from generated directed graph: $H_{G_{g}} = \text{GEDFN}(X, G_{g})$
    \State Concatenate embeddings: $H_{C} = [H_{G_{e}} \,\|\, H_{G_{g}}]$
    \State Feed $H_{C}$ into a DFN and apply final softmax layer to obtain predictions $\hat{Y}$
\end{algorithmic}
\end{algorithm}

\subsection{External and Generated Graph Construction}
To capture both complementary and hierarchical relationships among features, engGNN jointly utilizes an \textit{undirected external graph} and a \textit{directed generated graph}. The undirected graph encodes curated biological knowledge, while the directed graph captures task-specific dependencies inferred directly from data.

\begin{figure}[h!]
\centering
    \addtolength{\leftskip} {-2.6cm} 
    \addtolength{\rightskip}{-2.6cm} 
\includegraphics[width=\columnwidth]{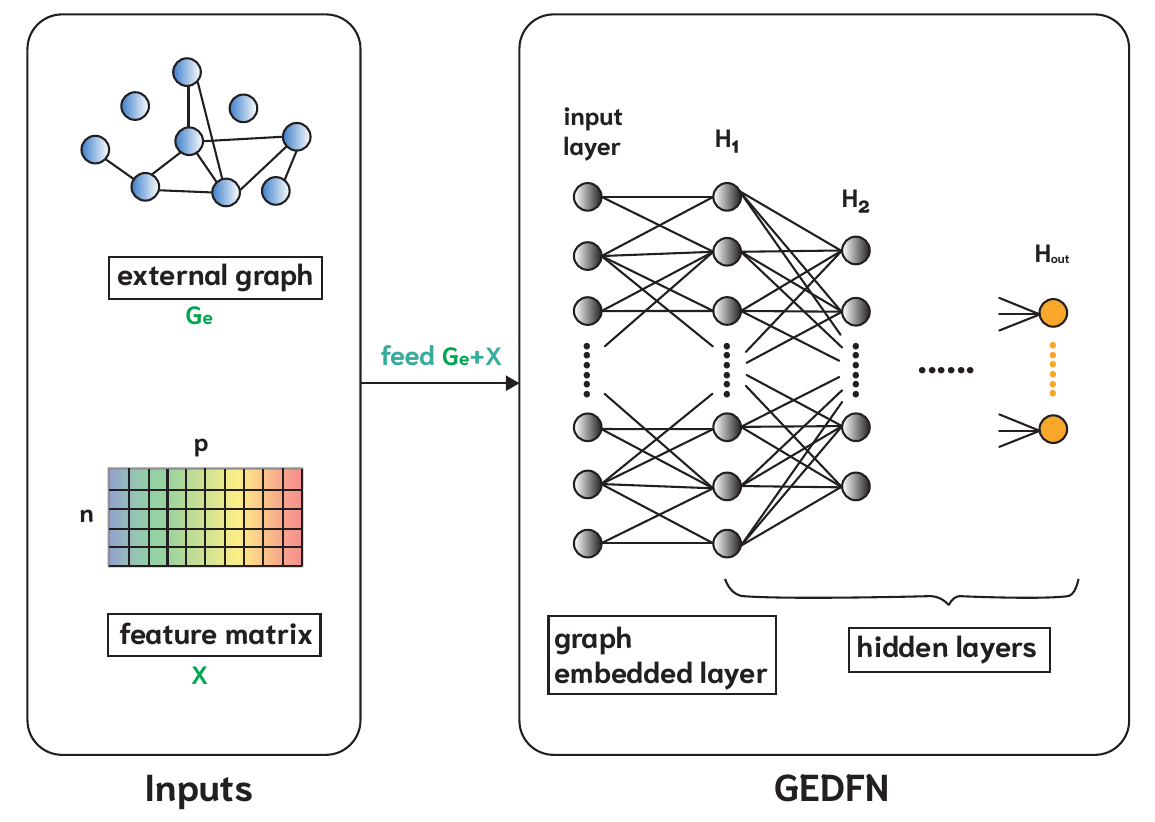}
\caption{Overview of $\mathrm{GEDFN}_{\text{e}}$. The left panel shows the inputs, including feature matrix $X$ and external undirected feature graph $G_e$, while the right panel shows the layer-wise structure of the graph-embedded deep feedforward network (GEDFN) model. The first hidden layer $H_1$ has the same number of neurons as the input feature size $p$. Connections from $X$ to $H_1$ reflect the adjacency structure of $G_e$ and are therefore symmetric, whereas subsequent layers (e.g., $H_1$ to $H_2$) are fully connected. The outputs are the last hidden layer $H_{out}$ of GEDFN. 
\label{GEDFN_e}}
\end{figure}

External graphs encode curated biological knowledge, where nodes represent molecular features (e.g., genes) and edges represent known functional relationships between them. In this study, we construct the external undirected feature graph $G_{e}$ using publicly available biological network databases to incorporate domain knowledge into the learning framework. 
This design enables engGNN to effectively leverage curated biological knowledge while remaining robust to incomplete or sparsely connected networks. Figure~\ref{GEDFN_e} provides the detailed structures of passing the external graph $G_e$ and feature matrix $X$ into the GEDFN model ($\mathrm{GEDFN}_{\text{e}}$).  

\begin{figure*}[h!]
\centering
    \addtolength{\leftskip} {-2.6cm} 
    \addtolength{\rightskip}{-2.6cm} 
\includegraphics[width=0.9\textwidth]{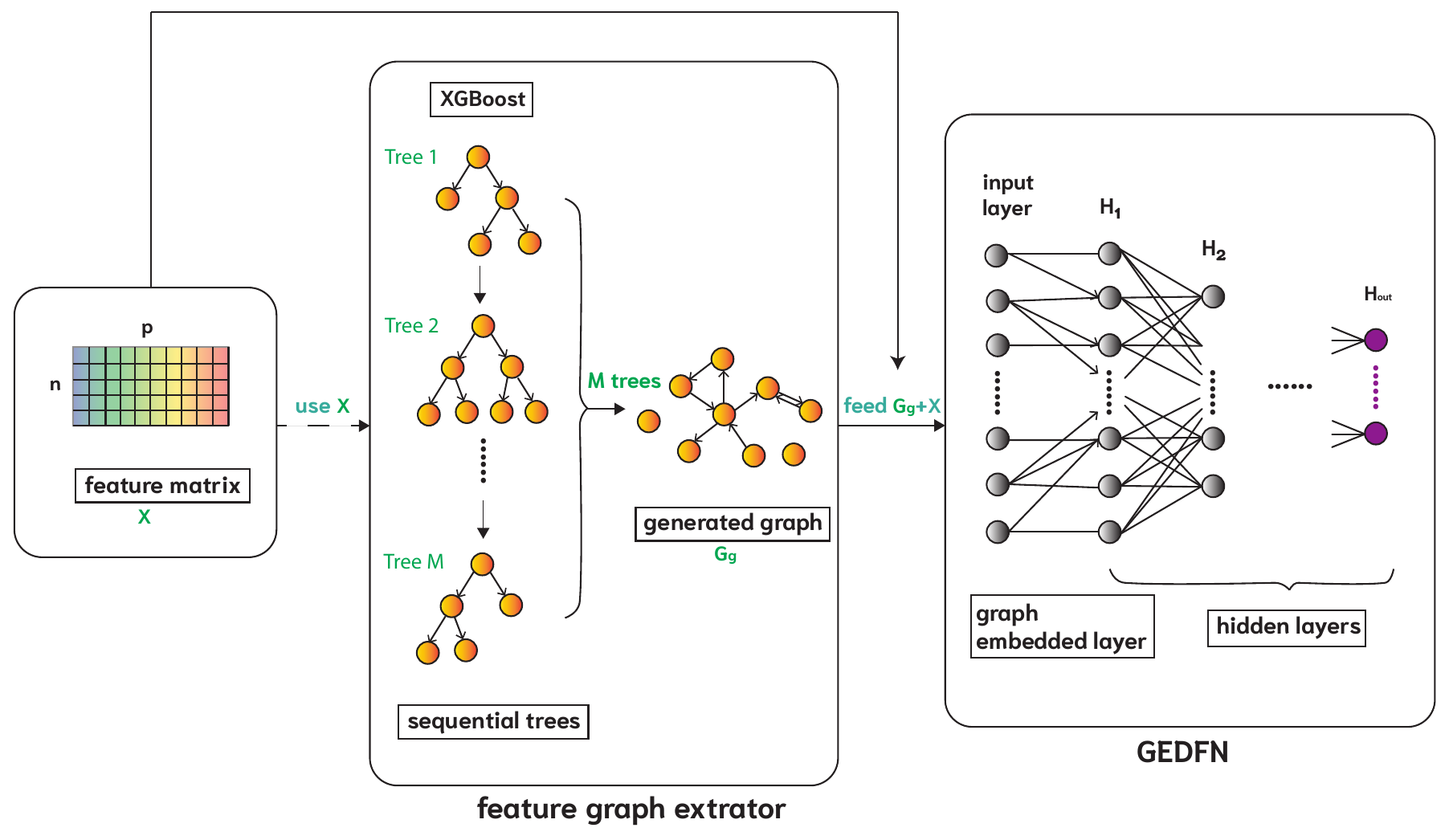}
\caption{Overview of $\mathrm{GEDFN}_{\text{g}}$, illustrated here using $\mathrm{GEDFN}_{\text{xgb}}$. The left panel illustrates the feature graph extractor using XGBoost, and the right panel presents the layer-wise structure of the graph-embedded deep feedforward network (GEDFN) model. The inputs are the feature matrix $X$ and the generated directed feature graph $G_g$. The first hidden layer $H_1$ has the same number of neurons as the input feature size $p$. Connections from $X$ to $H_1$ reflect the adjacency structure of $G_g$ and are therefore asymmetric, whereas subsequent layers (e.g., $H_1$ to $H_2$) are fully connected. The outputs are the last hidden layer $H_{out}$ of the GEDFN model.  
\label{GEDFN_g}}
\end{figure*}

To complement the external graph, we construct a directed feature graph tailored to the same prediction task using XGBoost. This ensures the generated graph reflects how features interact in contributing to the prediction, enabling engGNN to learn task-informed feature relationships rather than generic associations. We selected XGBoost over other tree-based methods, such as random forest \cite{breiman2001random} or gradient boosting machine (GBM) \cite{friedman2001greedy}, due to its sparsity-aware algorithm, efficient handling of sparse data with a weighted quantile sketch for approximate tree learning \cite{chen2016xgboost}, and built-in regularization mechanisms to mitigate the overfitting, an issue often observed in GBM. Following the forgeNet approach \cite{kong2020forgenet}, each decision tree defines a directed graph where nodes are selected features and edges reflect the hierarchical splitting order within the tree. For $M$ fitted decision trees, we constructed a corresponding set of directed graphs ${G_{m}(V_{m}, E_{m})}$, where $m = 1, 2, \ldots, M$. Each graph consists of a subset of selected features ($V_{m}$) and their directed connections ($E_{m}$) based on the tree structure. All $M$ graphs are merged to form an aggregated directed feature graph $G_g(V, E)$,
where $V = \bigcup_{m=1}^{M} V_{m}$ and $E = \bigcup_{m=1}^{M} E_{m}$.  
Its adjacency matrix is then augmented with self-loops and embedded into a GEDFN model for representation learning. Figure~\ref{GEDFN_g} illustrates the procedure for constructing the XGBoost-based graph $G_{g}$ and integrating it into GEDFN ($\mathrm{GEDFN}_{\text{xgb}}$). 

\subsection{Joint Graph Embedding via GNN} \label{section_GEDFN}
The core GNN component of engGNN is the graph-embedded deep feedforward network (GEDFN) \cite{kong2018graph}, which embeds a feature graph into the first hidden layer to enforce sparse and informative connections.  
The model starts with the original input features, represented by the feature matrix $X \in \mathbb{R}^{n \times p}$, where rows correspond to samples and columns correspond to measured molecular features (e.g., gene expression levels).
Unlike a standard deep feedforward network (DFN), which fully connects the input layer to the first hidden layer, GEDFN restricts connections based on the graph structure so that only neighboring features contribute to representation learning. This design encourages information sharing among biologically related features while reducing noise from unrelated ones. 
More specifically, GEDFN modifies the first linear transformation by incorporating the adjacency matrix of the feature graph. The embedding operation is defined as: 
$$H_{1} = \sigma(X(W\odot \tilde{A}) + b),$$ 
where $X$ is the input feature data matrix with $n$ samples and $p$ features, $ \tilde{A} \in \mathbb{R}^{p\times p}$ is the graph adjacency matrix (augmented with self-loops), $W$ as well as $b$ are the trainable weight matrix and bias, respectively, $\odot$ denotes the Hadamard (element-wise) product, and $\sigma(\cdot)$ is an activation function. Here, $\tilde{A}$ serves as a structural mask on $W$, ensuring that connections only exist between graph-adjacent features. By including self-loops, each transformed feature representation in the first hidden layer is computed as a weighted combination of the original feature and its neighbors. 
Subsequent hidden layers follow a standard DFN architecture and are fully connected.     
GEDFN is applied separately to the external graph $G_{e}$ and the generated graph $G_{g}$ to produce two graph-informed feature embeddings (denoted $H_{G_{e}}$ and $H_{G_{g}}$, respectively). These are then concatenated as $H_{C} = [H_{G_{e}} \,\|\, H_{G_{g}}]$ and passed to a DFN model for downstream representation learning and final classification. 

\subsection{Feature Importance and Interpretability} \label{IF}
To provide a reliable interpretation and support downstream biological discovery, we developed a feature importance (IF) scoring method by extending both the connection weight method \cite{olden2002illuminating} and the graph connection weight method \cite{kong2020forgenet, kong2018graph}. For each feature $j$, the overall importance score $IF_{j}$ is computed as the sum of two components: 
(1) $IF_{j}^{(G_{e})}$, the importance derived from the external graph $\mathrm{GEDFN}_{\text{e}}$, and 
(2) $IF_{j}^{(G_{g})}$, the importance derived from the generated graph $\mathrm{GEDFN}_{\text{g}}$.
Thus,
\begin{eqnarray}
IF_{j} = IF_{j}^{(G_{e})} + IF_{j}^{(G_{g})}. \label{eq:IF}
\end{eqnarray}
Each component is calculated as: 
\begin{eqnarray*} 
IF_{j}^{(G_{e})} &=& \sum_{u=1}^p \left|W_{ju}^{(in)(G_{e})}I(\tilde{A}_{ju}^{(G_{e})}=1)\right| \\
&+& \sum_{v=1}^p \left|W_{vj}^{(in)(G_{e})}I(\tilde{A}_{vj}^{(G_{e})}=1)\right|,
 \end{eqnarray*} 
\begin{eqnarray*} 
IF_{j}^{(G_{g})} &=& \sum_{u=1}^p \left|W_{ju}^{(in)(G_{g})}I(\tilde{A}_{ju}^{(G_{g})}=1)\right| \\
&+& \sum_{v=1}^p \left|W_{vj}^{(in)(G_{g})}I(\tilde{A}_{vj}^{(G_{g})}=1)\right|. 
 \end{eqnarray*} 
Here, $W^{(in)}$ are the weights connecting the input matrix $X$ to the first hidden layer $H_{1}$, and each element $W_{ju}^{(in)}$ corresponds to the entry in the $j$-th row and $u$-th column. Similarly, $\tilde{A}_{ju}$ represents the element in the $j$-th row and $u$-th column of the modified adjacency matrix $\tilde{A}$. The indicator function $I(\cdot)$ ensures that only connected nodes are considered in the summation. 
Thus, $IF_{j}^{(G_{e})}$ captures the undirected connection weights involving feature $j$ in the external graph $G_{e}$, while $IF_{j}^{(G_{g})}$ reflects the directed connections in the generated graph $G_{g}$. 
To make weights comparable across features, we apply $Z$-score normalization to the input matrix $X$ before model training, and the same normalized data were also used for computing feature importance scores. Although these scores lack direct intrinsic meaning, the relative ranking of features highlights the most influential ones and can guide downstream biological interpretation.

\subsection{Implementation Details} \label{implementation}

The implementation was built using TensorFlow \cite{abadi2016tensorflow} for the deep learning components and the XGBoost package \cite{chen2016xgboost} for graph generation. Neural networks were trained using the Adam optimizer \cite{kingma2014adam}, a variant of the stochastic gradient descent (SGD) algorithm. To improve computational efficiency and reduce memory load, we used mini-batch training with randomly shuffled subsets of samples. The Rectified Linear Unit (ReLU) activation function \cite{nair2010rectified}, defined as $\sigma(x) = max(0, x)$, was employed for nonlinearity. 
Model training minimized the categorical cross-entropy loss function: 
$$L(y, p) = -\sum_{i=1}^{K}y_{i}\ln p_i,$$
where $y = (y_{1},  \ldots, y_{K})$ represents the one-hot encoded true label vector and  $p = (p_1,  \ldots, p_K)$ denotes the predicted class probabilities obtained from the softmax output layer. 
Since our primary focus was on binary classification, we used $K=2$ softmax outputs to represent the two classes. 

For the XGBoost component, the number of trees (estimators) was set proportional to the feature size ($n_{\text{trees}} = 0.2p$ to balance model complexity and computational cost. For neural network components, we tuned hyperparameters including learning rate, batch size, dropout proportion, number of hidden layers (network depth), and number of neurons per layer (layer width). 
We performed a grid search over a feasible hyperparameter space and selected values that generalized well across datasets. Supplemental Table 1 summarizes this search space. 
For both $\mathrm{GEDFN}_{\text{e}}$ and $\mathrm{GEDFN}_{\text{g}}$, the width of the first hidden layer was fixed at the feature size $p$. Network depth was set to three, with the second and third layers containing 64 and 16 neurons, respectively. 
The DFN component had a single hidden layer with a width of 16, and the output layer used the softmax function to generate class probabilities. Other training settings were as follows: learning rate = 0.0001, number of epochs = 50, dropout rate = 0.2, and batch size = 16. The data were split into 80\% training and 20\% testing sets, and the runs were repeated 20 times. Early stopping was applied to minimize overfitting by stopping training when the loss failed to improve for several consecutive epochs. All experiments were conducted on Boston University's Shared Computing Cluster (SCC) using four CPUs per job.

\section{Simulation Experiments}
In this section, we generate multiple high-dimensional datasets under various scenarios to evaluate the performance of engGNN in both classification and feature selection tasks. 

\begin{table*}[h!]
\centering
\caption{Average classification accuracy, ROC-AUC, and F1-score of various models across nine simulation scenarios defined by $p_{n}$ (sample-to-feature ratio) and $p_{t}$ (true feature proportion). Best values in each row are shown in bold. Reported values are means (SD) over 20 runs.}
\label{tab:classification}
\begin{adjustbox}{max width=\textwidth}
\begin{tabular}{c|rr|rrrrrrr}
\toprule
Metric & $p_n$ & $p_t$ & engGNN & $\mathrm{GEDFN}_{\text{xgb}}$ & $\mathrm{GEDFN}_{\text{rf}}$ & $\mathrm{GEDFN}_{\text{e}}$ & DFN & XGBoost & RF \\
\midrule
\multirow{9}{*}{Accuracy} & 0.05 & 0.05 & \textbf{0.868 (0.0020)} & 0.859 (0.0021) & \textbf{0.868 (0.0020)} & 0.848 (0.0011) & 0.839 (0.0030) & 0.859 (0.0023) & 0.834 (0.0054) \\
& 0.05 & 0.10 & 0.870 (0.0029) & 0.860 (0.0022) & 0.877 (0.0016) & \textbf{0.886 (0.0019)} & 0.861 (0.0015) & 0.833 (0.0033) & 0.818 (0.0015) \\
& 0.05 & 0.20 & 0.905 (0.0010) & \textbf{0.919 (0.0012)} & 0.886 (0.0021) & 0.889 (0.0017) & 0.865 (0.0021) & 0.856 (0.0017) & 0.848 (0.0021) \\
& 0.10 & 0.05 & 0.908 (0.0006) & \textbf{0.910 (0.0006)} & 0.904 (0.0006) & 0.900 (0.0010) & 0.880 (0.0008) & 0.883 (0.0008) & 0.858 (0.0013) \\
& 0.10 & 0.10 & \textbf{0.912 (0.0009)} & 0.902 (0.0008) & 0.903 (0.0006) & 0.906 (0.0010) & 0.882 (0.0007) & 0.874 (0.0006) & 0.870 (0.0017) \\
& 0.10 & 0.20 & 0.908 (0.0007) & 0.910 (0.0003) & \textbf{0.914 (0.0004)} & 0.913 (0.0010) & 0.892 (0.0006) & 0.864 (0.0009) & 0.851 (0.0012) \\
& 0.20 & 0.05 & 0.923 (0.0002) & 0.923 (0.0004) & 0.920 (0.0003) & \textbf{0.924 (0.0002)} & 0.913 (0.0003) & 0.905 (0.0002) & 0.866 (0.0006) \\
& 0.20 & 0.10 & \textbf{0.926 (0.0003)} & 0.920 (0.0003) & 0.921 (0.0003) & 0.914 (0.0003) & 0.905 (0.0002) & 0.896 (0.0004) & 0.858 (0.0003) \\
& 0.20 & 0.20 & \textbf{0.909 (0.0003)} & 0.906 (0.0005) & 0.908 (0.0004) & 0.905 (0.0003) & 0.893 (0.0004) & 0.898 (0.0006) & 0.845 (0.0008) \\
\midrule
\multirow{9}{*}{ROC-AUC} & 0.05 & 0.05 & \textbf{0.952 (0.0010)} & \textbf{0.952 (0.0005)} & 0.949 (0.0009) & 0.943 (0.0005) & 0.937 (0.0007) & 0.947 (0.0008) & 0.926 (0.0016) \\
& 0.05 & 0.10 & 0.946 (0.0012) & 0.955 (0.0007) & \textbf{0.956 (0.0005)} & 0.951 (0.0006) & 0.932 (0.0012) & 0.921 (0.0010) & 0.909 (0.0012) \\
& 0.05 & 0.20 & 0.974 (0.0002) & \textbf{0.976 (0.0004)} & 0.959 (0.0009) & 0.965 (0.0004) & 0.943 (0.0010) & 0.935 (0.0011) & 0.927 (0.0011) \\
& 0.10 & 0.05 & \textbf{0.972 (0.0002)} & \textbf{0.972 (0.0001)} & 0.971 (0.0002) & 0.969 (0.0002) & 0.955 (0.0003) & 0.959 (0.0002) & 0.944 (0.0004) \\
& 0.10 & 0.10 & \textbf{0.968 (0.0002)} & 0.963 (0.0002) & 0.965 (0.0002) & 0.963 (0.0005) & 0.954 (0.0002) & 0.944 (0.0004) & 0.940 (0.0007) \\
& 0.10 & 0.20 & 0.971 (0.0001) & 0.975 (0.0001) & \textbf{0.976 (0.0001)} & 0.972 (0.0002) & 0.963 (0.0001) & 0.947 (0.0004) & 0.927 (0.0005) \\
& 0.20 & 0.05 & \textbf{0.981 (0.0000)} & \textbf{0.981 (0.0001)} & 0.978 (0.0001) & 0.980 (0.0000) & 0.974 (0.0001) & 0.969 (0.0001) & 0.942 (0.0002) \\
& 0.20 & 0.10 & \textbf{0.980 (0.0000)} & 0.978 (0.0001) & 0.978 (0.0001) & 0.975 (0.0001) & 0.970 (0.0001) & 0.964 (0.0001) & 0.942 (0.0002) \\
& 0.20 & 0.20 & \textbf{0.972 (0.0001)} & 0.971 (0.0001) & 0.968 (0.0001) & 0.970 (0.0001) & 0.961 (0.0002) & 0.963 (0.0002) & 0.930 (0.0003) \\
\midrule
\multirow{9}{*}{F1-score} & 0.05 & 0.05 & \textbf{0.833 (0.0036)} & 0.818 (0.0036) & 0.831 (0.0026) & 0.801 (0.0017) & 0.797 (0.0042) & 0.824 (0.0037) & 0.773 (0.0093) \\
& 0.05 & 0.10 & 0.838 (0.0041) & 0.834 (0.0026) & 0.836 (0.0023) & \textbf{0.847 (0.0025)} & 0.827 (0.0026) & 0.784 (0.0048) & 0.762 (0.0047) \\
& 0.05 & 0.20 & 0.881 (0.0020) & \textbf{0.894 (0.0023)} & 0.841 (0.0040) & 0.857 (0.0028) & 0.830 (0.0037) & 0.811 (0.0038) & 0.807 (0.0024) \\
& 0.10 & 0.05 & 0.885 (0.0009) & \textbf{0.886 (0.0010)} & 0.878 (0.0009) & 0.878 (0.0014) & 0.853 (0.0009) & 0.847 (0.0016) & 0.812 (0.0020) \\
& 0.10 & 0.10 & \textbf{0.885 (0.0013)} & 0.877 (0.0014) & 0.875 (0.0011) & 0.883 (0.0016) & 0.847 (0.0014) & 0.834 (0.0019) & 0.830 (0.0030) \\
& 0.10 & 0.20 & 0.879 (0.0014) & 0.882 (0.0008) & 0.886 (0.0011) & \textbf{0.889 (0.0020)} & 0.870 (0.0008) & 0.818 (0.0022) & 0.802 (0.0028) \\
& 0.20 & 0.05 & \textbf{0.903 (0.0003)} & 0.902 (0.0006) & 0.897 (0.0003) & \textbf{0.903 (0.0004)} & 0.888 (0.0007) & 0.880 (0.0005) & 0.826 (0.0010) \\
& 0.20 & 0.10 & \textbf{0.906 (0.0005)} & 0.898 (0.0006) & 0.904 (0.0005) & 0.890 (0.0006) & 0.879 (0.0005) & 0.866 (0.0006) & 0.817 (0.0005) \\
& 0.20 & 0.20 & \textbf{0.884 (0.0004)} & 0.880 (0.0009) & 0.882 (0.0005) & 0.878 (0.0006) & 0.865 (0.0006) & 0.867 (0.0012) & 0.796 (0.0012) \\
\bottomrule
\end{tabular}
\end{adjustbox}
\end{table*}
\begin{figure*}[h!]
\centering
    \addtolength{\leftskip} {-2.6cm} 
    \addtolength{\rightskip}{-2.6cm} 
\includegraphics[width=0.9\textwidth]{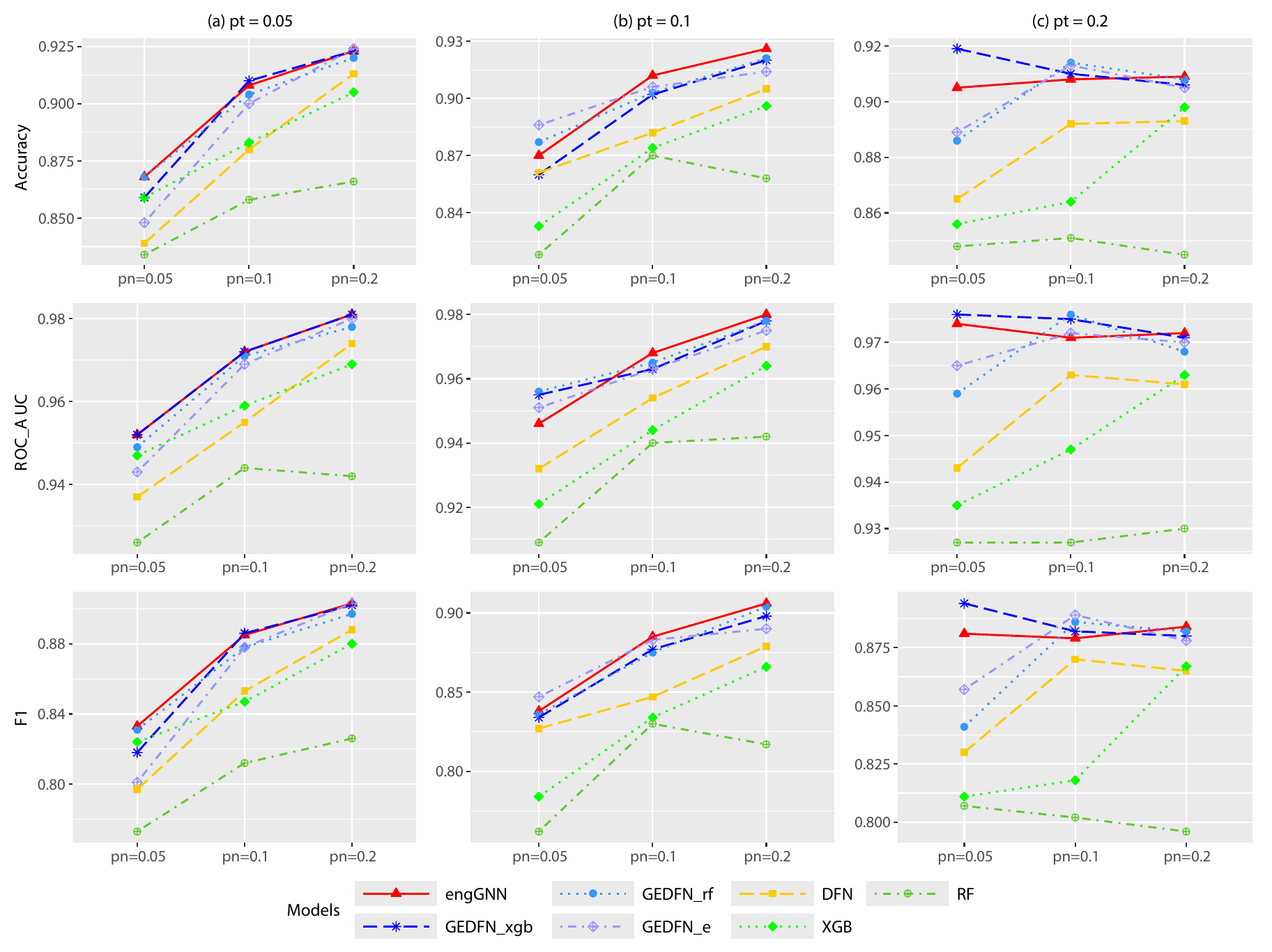}
\caption{Average classification accuracy, ROC-AUC, and F1-score of various models on simulated datasets with different sample-to-feature ratios ($p_{n} = 0.05, 0.1, 0.2$) and fixed proportions of true features ($p_{t} = 0.05, 0.1, 0.2$). The sample size is $n = 5000$, and each scenario was repeated 20 times. Plots (a) $p_{t}$ = 0.05; (b) $p_{t}$ = 0.1; (c) $p_{t}$ = 0.2.   \label{pt_fix_pn_accuracy_roc-auc_F1}}
\end{figure*}
\begin{figure*}[h!]
\centering
    \addtolength{\leftskip} {-2.6cm} 
    \addtolength{\rightskip}{-2.6cm} 
\includegraphics[width=0.9\textwidth]{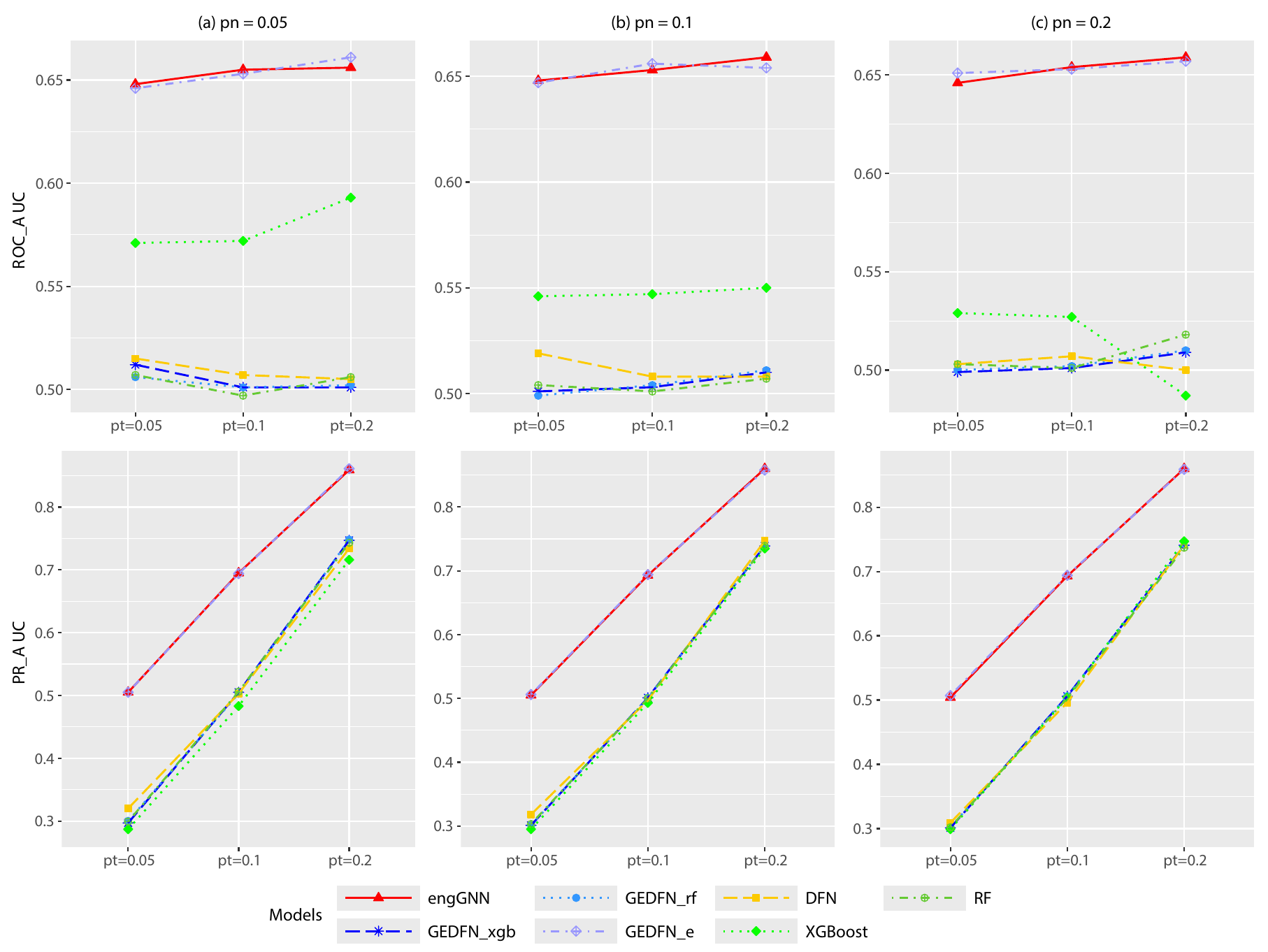}
\caption{Average feature selection ROC-AUC and PR-AUC of various models on simulated datasets across different values of proportion of true features ($p_{t} = 0.05, 0.1, 0.2$) with fixed sample size-to-feature size ratio ($p_{n}  = 0.05, 0.1, 0.2$). The sample size is $n = 5000$, and each scenario was repeated 20 times. Plots (a) $p_{t}$ = 0.05; (b) $p_{t}$ = 0.1; (c) $p_{t}$ = 0.2. Metrics include ROC-AUC and PR-AUC.  \label{pn_fix_roc-auc_pr-auc_FeatureSelection}}
\end{figure*}

\subsection{Synthetic Data Generation}
 
Each simulated dataset contained a binary outcome variable $Y$, a feature matrix $X \in \mathbb{R}^{n \times p}$, an external feature graph $G$, and its corresponding adjacency matrix $A \in \mathbb{R}^{p\times p}$. The external graph $G$ was designed to mimic biological connectivity patterns. Specifically, we created $G$ using the Barabási-Albert (B-A) model \cite{barabasi1999emergence}, which produces undirected scale-free networks with $p$ nodes. Such networks follow a power-law degree distribution to reflect the heterogeneous connectivity often observed in biological systems.
The corresponding adjacency matrix $A$ was then computed. 
To generate the feature matrix $X$, we created a weighted adjacency matrix $A^{\ast}$ by sampling edge weights from 0.1 to 10. Using $A^{\ast}$, we constructed the feature covariance matrix $\Sigma_{X}$ as: 
$$\Sigma_{X} = (I-A^{\ast})^{-1}\Sigma_{\epsilon}((I-A^{\ast})^{-1})^{T},$$
where $\Sigma_{\epsilon}$ is identity matrix. A random error term $\epsilon$ was drawn from a multivariate normal distribution with mean vector $\mu_{\epsilon}$. The mean vector for $X$ ($\mu_{X}$) was drawn uniformly from 7 to 13. Finally, the feature matrix was sampled as $X \sim MVN(\mu_{X}, \Sigma_{X})$.

The binary outcome $Y$ was generated using a subset of informative features. These features were selected based on their closeness centrality in $G$ \cite{beauchamp1965improved}. Nodes were stratified into groups based on centrality: (1) a high centrality group ($H_c$) with the top 50\% ranked nodes, and (2) a low centrality group ($L_c$) with the remaining nodes. To prioritize biologically relevant nodes, 80\% of true predictors were randomly selected from $H_c$ and 20\% from $L_c$. We expanded this set by including all one-hop neighbors, producing the final set of important predictors $X^{(\text{imp})}$.
We created $Y$ using a nonlinear transformation with a generalized linear model: 
$$y_{i}=I\left\{g\left(\beta_{0}+x_{i}^{(\text{imp})}\beta\right)>s\right\},$$ 
where $i = 1, \dots, n$, $I(\cdot)$ is the indicator function, $g(\cdot) = \exp(\phi(\cdot))+(\phi(\cdot))^{2}$ with $\phi(\cdot)$ denoting the min-max transformation, and the threshold $s = 0.6$ was set to ensure mild class imbalance. We generated the intercept term ($\beta_{0}$) from $N(-5, 5^2)$, and the feature coefficients $\beta$ were independently drawn from Uniform[-5, 5]. Random seeds were fixed for reproducibility, and the sample size was set at $n = 5000$. By varying the sample-to-feature ratio ($p_{n}$) and the proportion of true features ($p_{t}$) at 0.05, 0.1, and 0.2, we created datasets with feature sizes ($p$) of 250, 500, and 1000, and numbers of true features ($t$) of 250, 500, and 1000. This resulted in nine simulation scenarios that account for both the dimensionality and feature importance. The simulated data were divided into 80\% for training and 20\% for testing, with the procedure repeated across 20 independent replications.

\subsection{Classification Performance Evaluation} \label{Evaluation}
We compared engGNN with several state-of-the-art models, including (1) \textit{Ablation studies}: $\mathrm{GEDFN}_{\text{xgb}}$, $\mathrm{GEDFN}_{\text{rf}}$, and $\mathrm{GEDFN}_{\text{e}}$; and (2) \textit{Classical deep learning and machine learning methods}: DFN, XGBoost, and RF. Among these, $\mathrm{GEDFN}_{\text{e}}$ uses only the external graph (representing prior biological connections), while $\mathrm{GEDFN}_{\text{xgb}}$ and $\mathrm{GEDFN}_{\text{rf}}$ use feature graphs generated by XGBoost and RF, respectively. This comparison helps assess the contribution of external biological priors and data-driven graph construction to model performance. 
We evaluated classification performance using accuracy, the area under the receiver operating characteristic curve (ROC-AUC), and F1-score. 
Accuracy measures the overall proportion of correctly classified samples and is defined as
$$\mathrm{Accuracy} = \frac{\text{TP} + \text{TN}}{\text{TP + TN + FP + FN}},$$
The F1-score, defined as 
$$\frac{2\times \text{Precision} \times \text{Recall}}{\text{Precision} + \text{Recall}},$$
is the harmonic mean of precision and recall, where  
$$\mathrm{Precision} = \frac{\text{TP}}{\text{TP + FP}}, \quad \mathrm{Recall} = \frac{\text{TP}}{\text{TP + FN}},$$
and TP, TN, FP, and FN denote the numbers of true positives, true negatives, false positives, and false negatives, respectively. All metrics range from 0 to 1, with higher values indicating better performance. The reported classification results were averaged across the 20 replications. 

Table~\ref{tab:classification} highlights the dominant performance of our proposed model, engGNN, across accuracy, ROC-AUC, and F1-score in all nine scenarios defined by the sample-to-feature ratio ($p_{n}$) and the true feature proportion ($p_{t}$), each evaluated over 20 runs. When $p_{t} = 0.05$, engGNN consistently outperformed all other models across all $p_{n}$ values. At $p_{t} = 0.1$, engGNN remained dominant in most scenarios, particularly when $p_{n} = 0.1$ and $p_{n} = 0.2$. In the more challenging setting of $p_{n} = 0.05$, its performance (accuracy = 0.870, ROC-AUC = 0.946, and F1-score = 0.838) was slightly lower than that of $\mathrm{GEDFN}_{\text{e}}$ (accuracy = 0.886, ROC-AUC = 0.956, and F1-score = 0.847). For $p_{t} = 0.2$, engGNN remained among the top-performing models, with performance closely matching that of $\mathrm{GEDFN}_{\text{xgb}}$. In summary, for accuracy, engGNN achieved the best or tied performance in 4 out of 9 scenarios. For ROC-AUC, engGNN led in 6 out of 9 scenarios. For the F1-score, engGNN ranked best or tied in 5 out of 9 scenarios. Beyond engGNN’s consistent dominance, the results also reveal notable but limited strengths of other baselines. $\mathrm{GEDFN}_{\text{xgb}}$ was the most competitive alternative, with 4 wins in ROC-AUC; however, its advantage did not extend to accuracy or F1-score. $\mathrm{GEDFN}_{\text{rf}}$ achieved two wins each in accuracy and ROC-AUC, but failed to excel in F1-score. $\mathrm{GEDFN}_{\text{e}}$, which relies on a true external graph, won 3 times on F1-score, yet underperformed in ROC-AUC, indicating limited generalizability. In contrast, simpler methods (DFN, XGBoost, RF) rarely achieved top performance, reinforcing their limited effectiveness in high-dimensional, feature-sparse settings.

Figure~\ref{pt_fix_pn_accuracy_roc-auc_F1} further illustrates these findings. Overall, performance improves with increasing values of $p_{n}$ and $p_{t}$. Although individual baseline models occasionally performed well, only engGNN demonstrated consistent superiority across all evaluation metrics and settings. 
Supplemental Figure 1 summarizes the average classification performance across different $p_{n}$ or $p_{t}$ values, reaffirming engGNN's consistent advantage. Supplemental Figure 2 visualizes the average accuracy, ROC-AUC, and F1-score across all nine scenarios, while Supplemental Figure 3 further confirms engGNN’s superior overall classification performance. Together, these results highlight not only engGNN's robust and consistent classification performance across diverse scenarios but also its improved stability, as evidenced by lower variance across runs compared to competing models.

\subsection{Feature Selection Evaluation}

Since the true set of important features is available in the simulated datasets, we directly evaluated the performance of each feature selection method. A binary label was assigned to each feature: 1 if it was important and 0 otherwise. For $\mathrm{GEDFN}_{\text{xgb}}$, $\mathrm{GEDFN}_{\text{rf}}$, $\mathrm{GEDFN}_{\text{e}}$, and DFN, feature importance scores were computed using a similar procedure described in Section \ref{IF}. For XGBoost, we used the Gain metric (the average improvement in loss when a feature was used for splitting); for RF, we used the Gini importance measure (mean decrease in impurity). To ensure comparability across models, all importance scores were transformed into pseudo-probabilities using percentile ranks. Evaluation metrics include ROC-AUC and PR-AUC (Area under the Precision-Recall curve) \cite{richardson2006markov, yue2007support}, both of which range from 0 to 1, with higher values indicating better feature selection performance. Reported feature selection results represent the average across 20 independent runs. 

Figure \ref{pn_fix_roc-auc_pr-auc_FeatureSelection} and Supplementary Table 2 present results across varying proportions of true features ($p_{t} = 0.05, 0.1, 0.2$) under fixed sample-to-feature ratios ($p_{n}  = 0.05, 0.1, 0.2$). Across all conditions, engGNN consistently achieved the strongest feature selection performance, yielding the highest ROC-AUC and PR-AUC. For example, when $p_{t} = 0.1$ and $p_{n} = 0.2$, engGNN attained a ROC-AUC of 0.654 and PR-AUC of 0.693, compared to 0.653 and 0.694 for $\mathrm{GEDFN}_{\text{e}}$. All other baselines stayed near 0.50, approximately random performance. 
$\mathrm{GEDFN}_{\text{e}}$ performed competitively, particularly under lower $p_{t}$ values. However, as shown in Table \ref{tab:classification}, its classification accuracy was consistently lower than that of engGNN, highlighting that engGNN achieves a stronger balance between predictive power and interpretability. 
For ROC-AUC, both engGNN and $\mathrm{GEDFN}_{\text{e}}$ clearly outperformed all other models, with their advantage becoming more apparent as $p_{n}$ increased. 
For PR-AUC, the separation was even clearer: engGNN achieved large gains over competing methods, particularly in high-signal settings such as $p_{t} = 0.2$. For example, when $p_{t} = 0.2$ and $p_{n} = 0.2$, engGNN achieved a PR-AUC of 0.860, slightly above $\mathrm{GEDFN}_{\text{e}}$ at 0.859, while $\mathrm{GEDFN}_{\text{xgb}}$ and RF remained far behind at 0.741 and 0.737, respectively. 
In contrast, $\mathrm{GEDFN}_{\text{xgb}}$, $\mathrm{GEDFN}_{\text{rf}}$, DFN, XGBoost, and RF showed only marginal improvements across scenarios, with PR-AUC values often close to random (around 0.30-0.50) in low-signal settings such as $p_{t} = 0.05$ and $p_{t} = 0.1$.

Supplementary Figures 6-8 further reinforce these findings. Supplemental Figure 4 displays the average ROC-AUC and PR-AUC across varying $p_{t}$ values, showing steady improvement with higher $p_{t}$, especially for PR-AUC. Supplemental Figure 5 presents results across all nine simulation scenarios, with engGNN dominating nearly every case. Supplemental Figure 6 provides an integrated view, confirming once again that engGNN's superior performance in feature selection. 
Together, these results demonstrate that engGNN not only delivers leading classification performance but also provides robust and stable feature selection, making it uniquely effective in high-dimensional, feature-sparse applications.

\begin{table}[ht]
\caption{Classification performance of various models on real Alzheimer's disease (AD) gene expression data. Each scenario was replicated 20 times. Results are reported as mean (standard deviation). Best-performing values are bolded. Superscripts $^{*}$ indicate that engGNN is statistically significantly better than the corresponding baseline (p-value $<$ 0.001) based on Welch's $t$-test for unequal variances. } 
\label{RealData_AD_accuracy_roc-auc_F1}
\centering
\begin{adjustbox}{max width=\columnwidth}
\begin{tabular}{lrrr}
\toprule
\textbf{Model} & \textbf{Accuracy} & \textbf{ROC-AUC} & \textbf{F1-score} \\
\midrule
engGNN & \textbf{0.788 (0.0007)} & \textbf{0.855 (0.0007)} & \textbf{0.579 (0.0054)} \\
$\mathrm{GEDFN}_{\text{xgb}}$ & 0.777 (0.0012)$^{*}$ & 0.852 (0.0010)$^{*}$ & 0.571 (0.0036)$^{*}$ \\
$\mathrm{GEDFN}_{\text{rf}}$ & 0.771 (0.0013)$^{*}$ & 0.849 (0.0009)$^{*}$ & 0.565 (0.0046)$^{*}$ \\
$\mathrm{GEDFN}_{\text{e}}$ & 0.758 (0.0012)$^{*}$ & 0.841 (0.0008)$^{*}$ & 0.535 (0.0070)$^{*}$ \\
DFN & 0.766 (0.0008)$^{*}$ & 0.843 (0.0005)$^{*}$ & 0.575 (0.0068)$^{*}$ \\
XGB & 0.737 (0.0012)$^{*}$ & 0.809 (0.0009)$^{*}$ & 0.503 (0.0026)$^{*}$ \\
RF & 0.765 (0.0012)$^{*}$ & 0.846 (0.0011)$^{*}$ & 0.531 (0.0025)$^{*}$ \\
\bottomrule
\end{tabular}
\end{adjustbox}
\end{table}
\begin{table*}
\caption{Top 10 significantly enriched KEGG pathways based on the top 1000 genes prioritized by engGNN in the AD dataset. Enrichment was assessed using KEGG over-representation analysis. Columns: GeneRatio (the number of genes within that list that are annotated to the gene set divided by the size of the list of genes of interest), BgRatio (the number of genes within that distribution that are annotated to the gene set of interest divided by the total number of genes in the background distribution), p-value (raw), p.adjust (Benjamini-Hochberg adjusted), and q-value (FDR estimate). Well-documented AD pathways are shown in bold. } \label{pathways}
\begin{center}
\begin{tabular}{llrrrrr}
\toprule
\textbf{ID} & \textbf{Description} & \textbf{GeneRatio} & \textbf{BgRatio} & \textbf{p-value} & \textbf{p.adjust} & \textbf{q-value} \\ 
\midrule
hsa04020 & \textbf{Calcium signaling pathway} & 29/354 & 240/8191 & 4.33E-07 & 0.00013 & 9.30E-05 \\ 
hsa04550 & Signaling pathways regulating pluripotency of stem cells & 20/354 & 143/8191 & 3.01E-06 & 0.00033 & 0.00024 \\ 
hsa04010 & \textbf{MAPK signaling pathway} & 31/354 & 294/8191 & 3.37E-06 & 0.00033 & 0.00024 \\ 
hsa04925 & Aldosterone synthesis and secretion & 15/354 & 98/8191 & 1.76E-05 & 0.0013 & 0.00095 \\ 
hsa04024 & \textbf{cAMP signaling pathway} & 24/354 & 221/8191 & 2.74E-05 & 0.0014 & 0.0010 \\ 
hsa01521 & EGFR tyrosine kinase inhibitor resistance & 13/354 & 79/8191 & 2.92E-05 & 0.0014 & 0.0010 \\ 
hsa04015 & Rap1 signaling pathway & 23/354 & 210/8191 & 3.52E-05 & 0.0015 & 0.0011 \\ 
hsa04114 & Oocyte meiosis & 17/354 & 131/8191 & 4.50E-05 & 0.0017 & 0.0012 \\ 
hsa04911 & Insulin secretion & 13/354 & 86/8191 & 7.32E-05 & 0.0022 & 0.0016 \\ 
hsa01522 & Endocrine resistance & 14/354 & 98/8191 & 7.35E-05 & 0.0022 & 0.0016 \\
\bottomrule
\end{tabular}
\end{center}
\end{table*}

\section{Real Data Applications}

We applied the engGNN model to a real-world peripheral blood gene expression dataset for Alzheimer's Disease (AD) \cite{ad2019paper}. AD is a progressive neurodegenerative disorder characterized by cognitive decline and behavioral impairments. The goal of this analysis was to classify AD cases versus controls and to prioritize genes associated with the disease. 

\noindent \textbf{Data Processing and Graph Construction.} We retrieved gene expression profiles from the National Library of Medicine (NCBI) Gene Expression Omnibus (GEO) database (accession: GSE140831) \cite{GSE140831, nachun2019systems}, which includes samples from 204 AD patients and 530 cognitively normal controls. 
We standardized the gene expression values using the $Z$-score transformation before model training.
To incorporate biological prior knowledge, we obtained a brain tissue-specific functional gene network from 
HumanBase (GIANT) \cite{greene2015understanding}. This network served as the external undirected feature graph ($G_{e}$) for engGNN, where nodes represent genes and edges represent known functional interactions between them. Only edges with confirmed tissue-specific interactions were included to ensure biological relevance. We harmonized gene identifiers by mapping ENTREZ IDs to gene symbols with the R/Bioconductor \textit{biomaRt} package \cite{biomart}. We excluded genes without gene expression profiles and retained isolated nodes to maintain full feature coverage. The resulting external network contained 22,372 nodes (genes) and 12,116,734 edges. This external graph was then integrated with the data-driven, tree-generated graph ($G_{g}$) within engGNN for classification and feature selection. 

\noindent \textbf{Classification Results.} The real data was partitioned into 80\% for training and 20\% for testing, and the analysis was repeated across 20 independent runs. Further implementation details are provided in Section~\ref{implementation}. Table~\ref{RealData_AD_accuracy_roc-auc_F1} compares the average classification performance of engGNN with several baselines on the processed AD dataset. Hyperparameters were tuned following the same procedure as in the simulation study. engGNN achieved the best performance across all metrics, achieving the highest accuracy (0.788), ROC-AUC (0.855), and F1-score (0.579), with statistically significant improvements (p-value $<$ 0.001) over every baseline, as determined by Welch's $t$-test for unequal variances. 
Among the baselines, $\mathrm{GEDFN}_{\text{xgb}}$ performed relatively well on ROC-AUC (0.852) and accuracy (0.777). $\mathrm{GEDFN}_{\text{rf}}$ showed moderate results but remained below engGNN on every metric. $\mathrm{GEDFN}_{\text{e}}$ performed worse despite leveraging the external graph. Simpler models (DFN, XGBoost, RF) consistently underperformed, with weaker accuracy, ROC-AUC, and F1-scores.
Overall, only engGNN demonstrated consistently superior results across all evaluation metrics, confirming its robustness, stability, and practical utility for real-world biological data.

\noindent \textbf{Pathway Enrichment Analysis.} To evaluate the biological interpretability of the genes prioritized by engGNN, we performed KEGG pathway over-representation analysis using the R/Bioconductor \textit{clusterProfiler} package \cite{wu2021clusterprofiler}. Genes were ranked by their importance scores, and the top 1000 were selected for analysis (the full list is provided in the supplemental materials). 
Table \ref{pathways} shows the top 10 significantly enriched KEGG pathways. The calcium signaling pathway was the most enriched. This finding aligns with numerous studies that disruption of calcium homeostasis can play a critical role in the pathogenesis of AD, as altered neural calcium signaling could induce synaptic deficits and promote the accumulation of Amyloid beta plaques and neurofibrillary tangles \cite{laferla2002calcium, tong2018calcium, berridge2011calcium, berridge2013dysregulation}. Other top-ranked pathways, including MAPK signaling and cAMP signaling, are also well documented in the AD literature \cite{johnson2003p38, kim2010pathological, du2019mkp, chen2012alzheimer}. Supplemental Figures 7-9 provide detailed visualizations of these three pathways, highlighting genes from the top 1000 genes prioritized by engGNN in red. 
In addition, several less-studied pathways, such as ``signaling pathways regulating pluripotency of stem cells" and ``endocrine resistance", were also significantly enriched. While these pathways are not directly linked to AD in the existing literature, their emergence underscores engGNN's capacity to uncover potentially novel or underexplored biological mechanisms that may warrant further investigation.

Overall, engGNN not only achieved state-of-the-art classification performance but also successfully prioritized biologically meaningful genes. The enriched pathways align well with established AD mechanisms while also pointing to novel biology hypotheses. These findings support engGNN’s dual strengths in both predictive modeling and biological interpretability, which reinforces its potential for real-world biomedical applications.

\section{Conclusion and Discussion}
In this study, we proposed engGNN, a dual-graph neural network framework that integrates both directed and undirected graph structures for disease classification and feature selection in high-dimensional omics data. By integrating external biological knowledge with data-driven graph construction, engGNN provides a robust, interpretable, and biologically relevant approach to biomedical prediction tasks. Across both simulated datasets and real-world Alzheimer's Disease gene expression data, engGNN consistently outperformed or matched strong baselines. Its stability across diverse scenarios underscores its reliability for biomedical applications, where reproducibility and interpretability are as critical as predictive accuracy.

The main contributions of this work are threefold. First, engGNN jointly leverages both curated external and tree-generated feature graphs, producing richer and more informative embeddings that improve classification performance. Second, it yields interpretable feature importance scores, which enable downstream analyses such as pathway enrichment to reveal biologically meaningful mechanisms. Third, its modular and flexible design allows for seamless integration of additional graph sources or straightforward adaptation to other omics domains. While this study primarily focuses on binary disease classification with single-omics data, the framework lays a solid foundation for several promising extensions: (1) extending engGNN to multi-class disease prediction would broaden its utility, particularly for diseases with multiple subtypes; (2) incorporating multi-omics data, such as methylomics, proteomics, or metabolomics, could further enhance biological insight by integrating complementary layers of molecular information. 

In summary, engGNN represents a practical and interpretable framework that strikes a balance between predictive performance and biological relevance. It serves as a valuable tool for disease classification and biomarker discovery, and provides a strong foundation for future advancements in graph-based biomedical modeling.

\section*{Data and Code Availability}
The original gene expression data were obtained from the Gene Expression Omnibus (GEO) database (GSE140831 dataset: \url{https://www.ncbi.nlm.nih.gov/geo/query/acc.cgi?acc=GSE140831}). The original gene network was obtained from HumanBase/Tissue-specific gene networks (GIANT: \url{https://hb.flatironinstitute.org/download}). 
The preprocessed real data and the engGNN implementation code will be released publicly upon publication.

\begin{acks}
National Institutes of Health (NIH) R01DK122503 and National Institute of General Medical Sciences of the National Institutes of Health (NIH/NIGMS) P20GM104420 partially supported this work.
\end{acks}


\bibliographystyle{main}
\bibliography{main}

\appendix

\setcounter{figure}{0}
\setcounter{table}{0}

\renewcommand{\thefigure}{\arabic{figure}}
\renewcommand{\thetable}{\arabic{table}}

\renewcommand{\figurename}{\textbf{Supplemental Figure}}
\renewcommand{\tablename}{\textbf{Supplemental Table}}

\section{Supplemental Material}



\begin{figure*}[h]
\centering
    \addtolength{\leftskip} {-2.6cm} 
    \addtolength{\rightskip}{-2.6cm} 
\includegraphics[width=\textwidth]{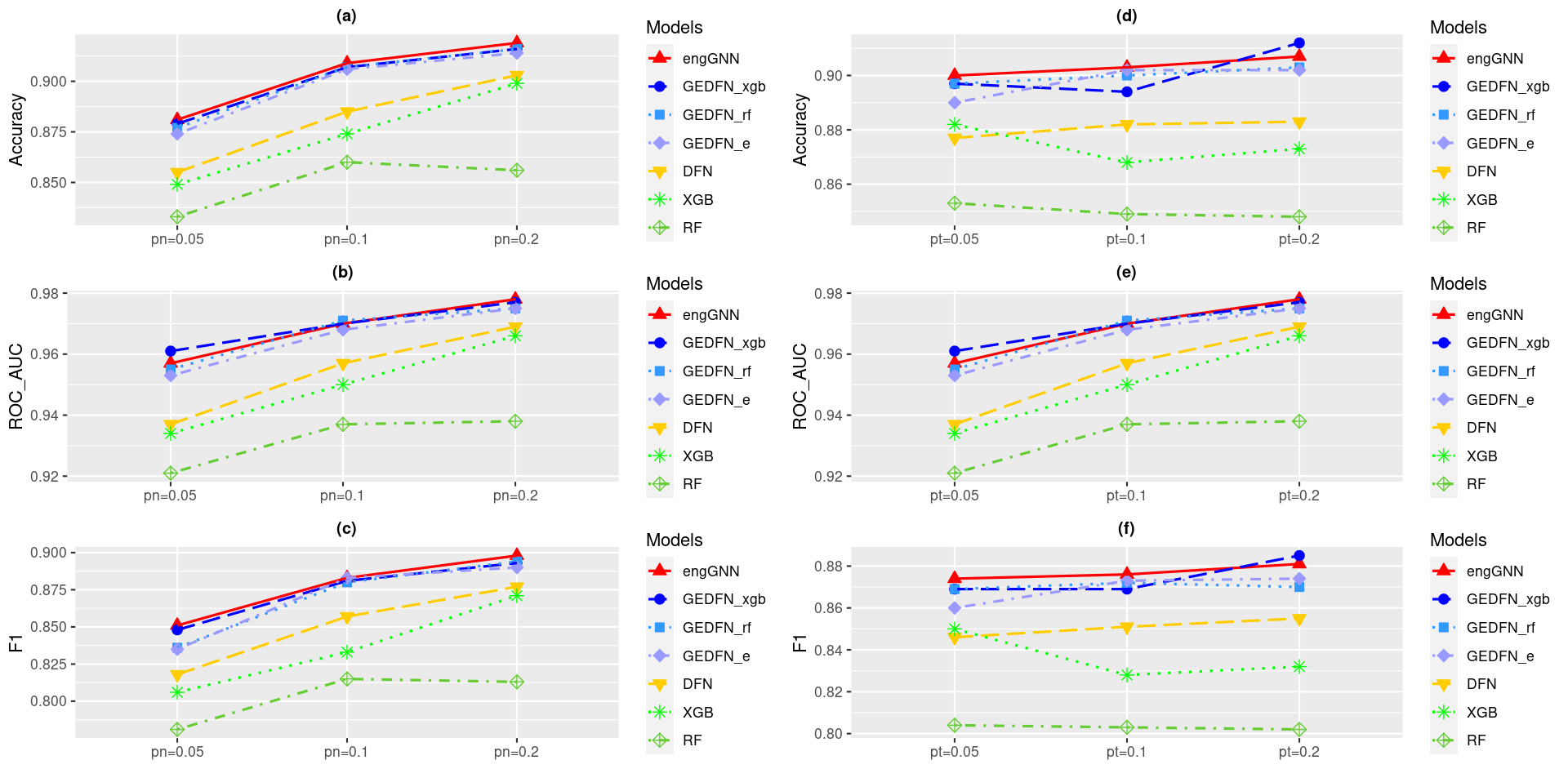}
\caption{Average classification performance of various models on simulated datasets across different values of sample size-to-feature size ratio ($p_{n}$) and proportion of true features ($p_{t}$). $p_{n}, p_{t} \in \{0.05, 0.1, 0.2\}$. Metrics include accuracy, ROC-AUC, and F1-score. Subplots (a) for different $p_{n}$ values; Subplots (b) for different $p_{t}$ values.  \label{pt_pn_accuracy_roc-auc_F1}}
\end{figure*}
\clearpage

\begin{figure*}[h]
\centering
    \addtolength{\leftskip} {-2.6cm} 
    \addtolength{\rightskip}{-2.6cm} 
\includegraphics[width=0.8\textwidth]{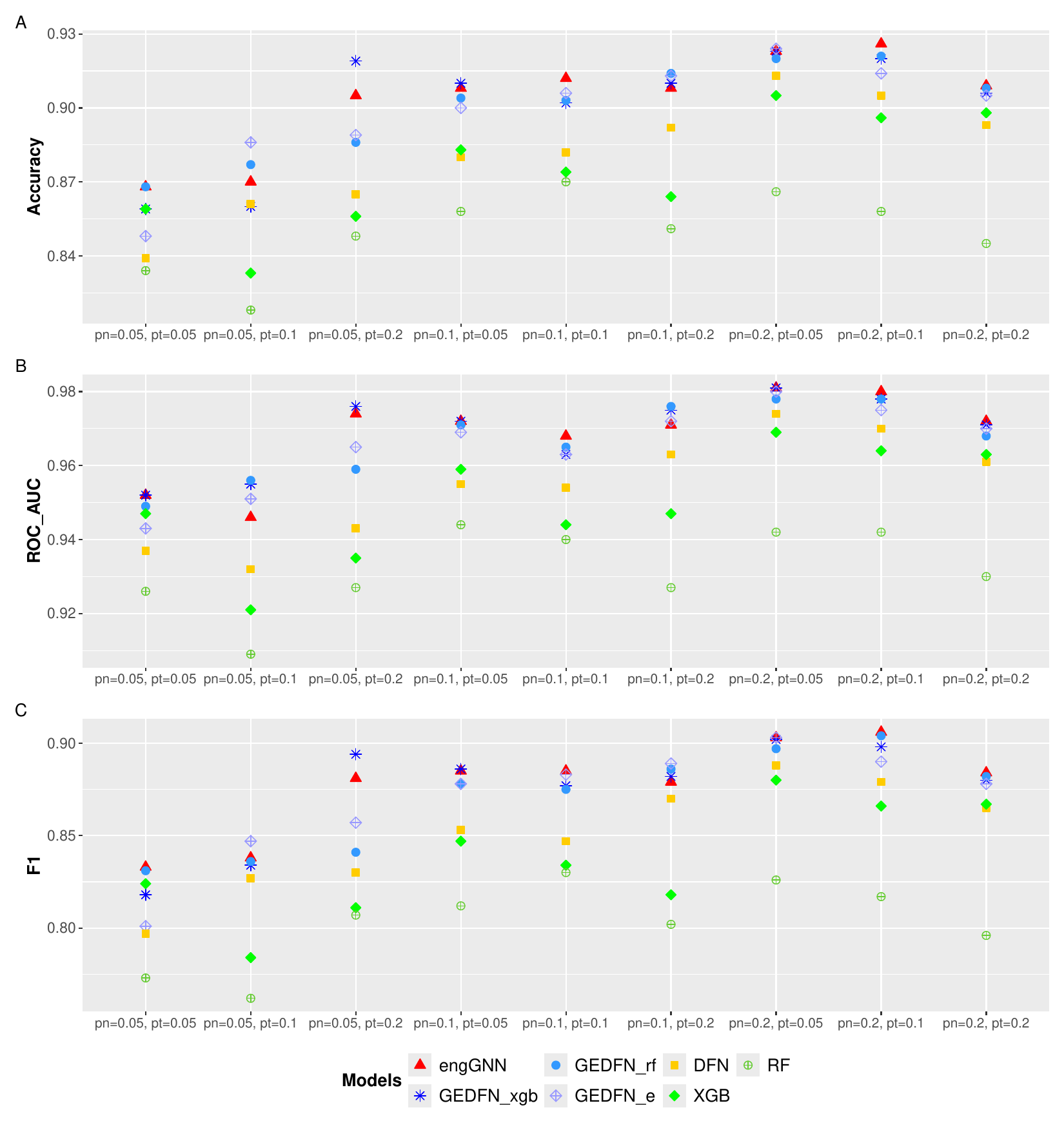}
\caption{Average classification performance of various models on simulated datasets across nine scenarios. The sample size is $n = 5000$, and each scenario was replicated 20 times. \label{CombinedAccuracy_ROCAUC_F1_9scenarios}}
\end{figure*}
\clearpage

\begin{figure*}[h]
\centering
    \addtolength{\leftskip} {-2.6cm} 
    \addtolength{\rightskip}{-2.6cm} 
\includegraphics[width=0.8\textwidth]{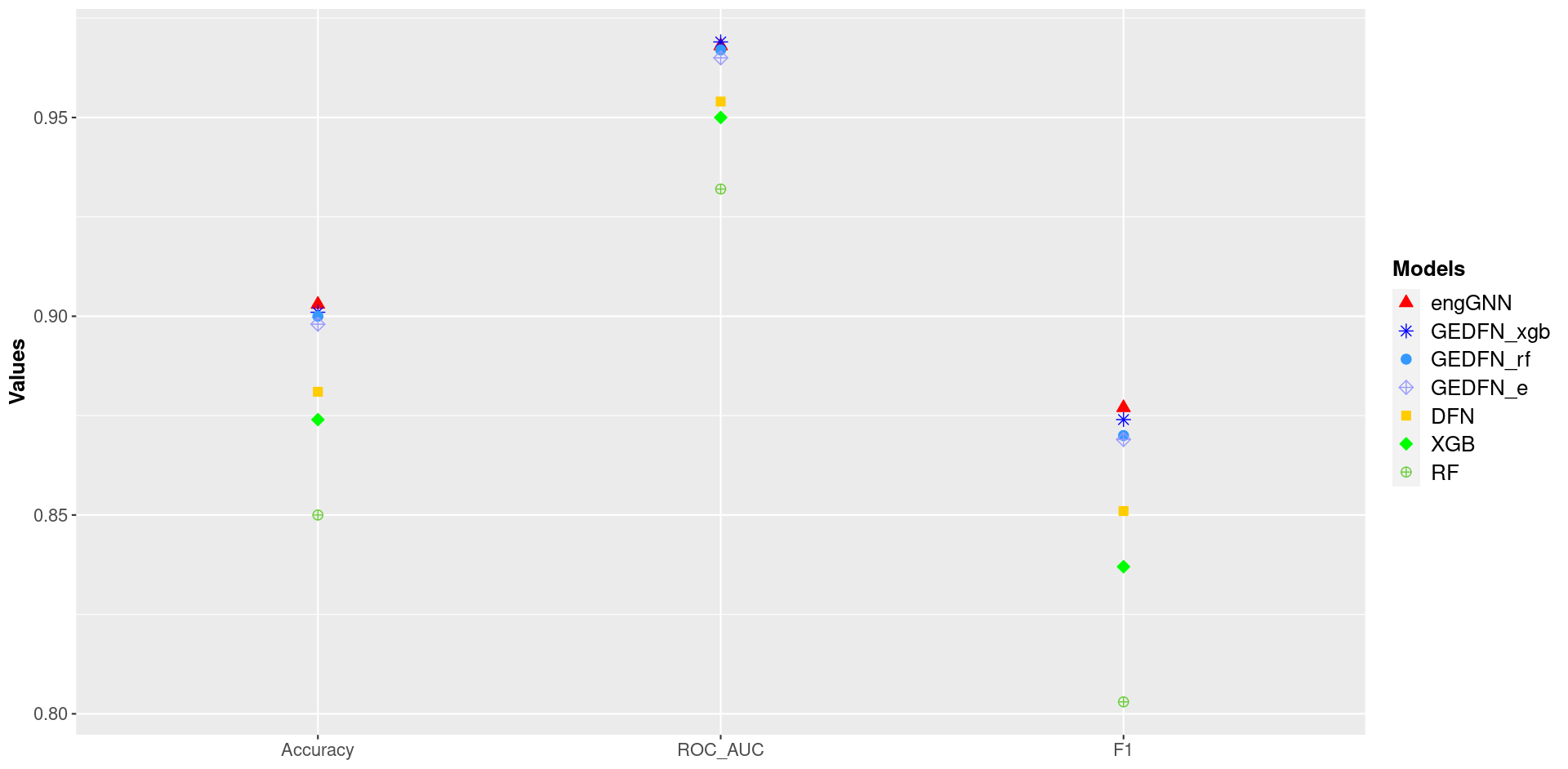}
\caption{Average classification accuracy, ROC-AUC, and F1-score of various models on simulated datasets across all the scenarios. \label{all_accuracy_roc-auc_F1}}
\end{figure*}
\clearpage

\begin{figure*}[h]
\centering
    \addtolength{\leftskip} {-2.6cm} 
    \addtolength{\rightskip}{-2.6cm} 
\includegraphics[width=0.8\textwidth]{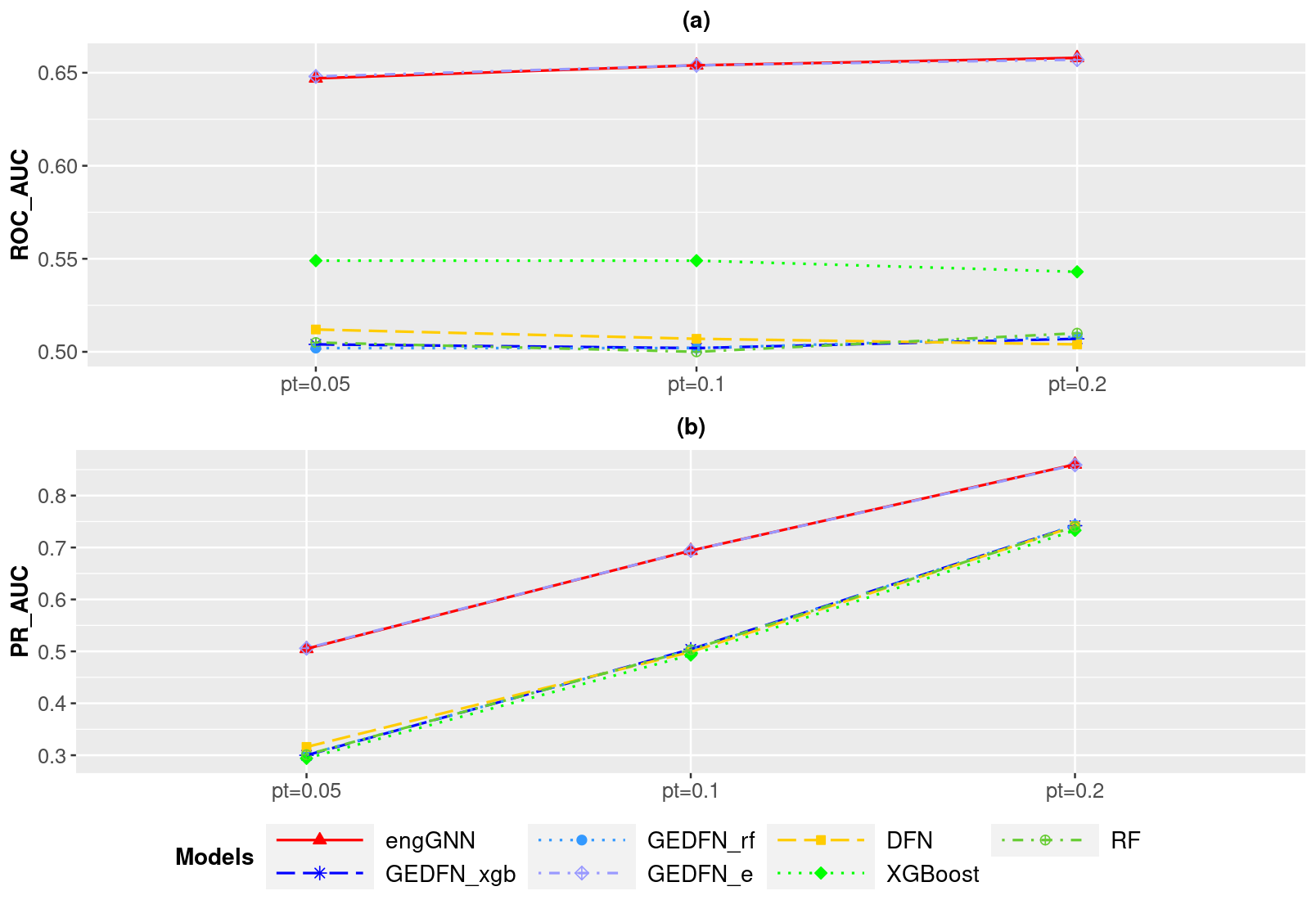}
\caption{Average feature selection performance of various models on simulated datasets across different proportions of true features ($p_{t} = 0.05, 0.1, 0.2$). Plot (a) ROC-AUC; Plot (b) PR-AUC.  \label{pt_roc-auc_pr-auc_FeatureSelection}}
\end{figure*}
\clearpage

\begin{figure*}[h]
\centering
    \addtolength{\leftskip} {-2.6cm} 
    \addtolength{\rightskip}{-2.6cm} 
\includegraphics[width=0.8\textwidth]{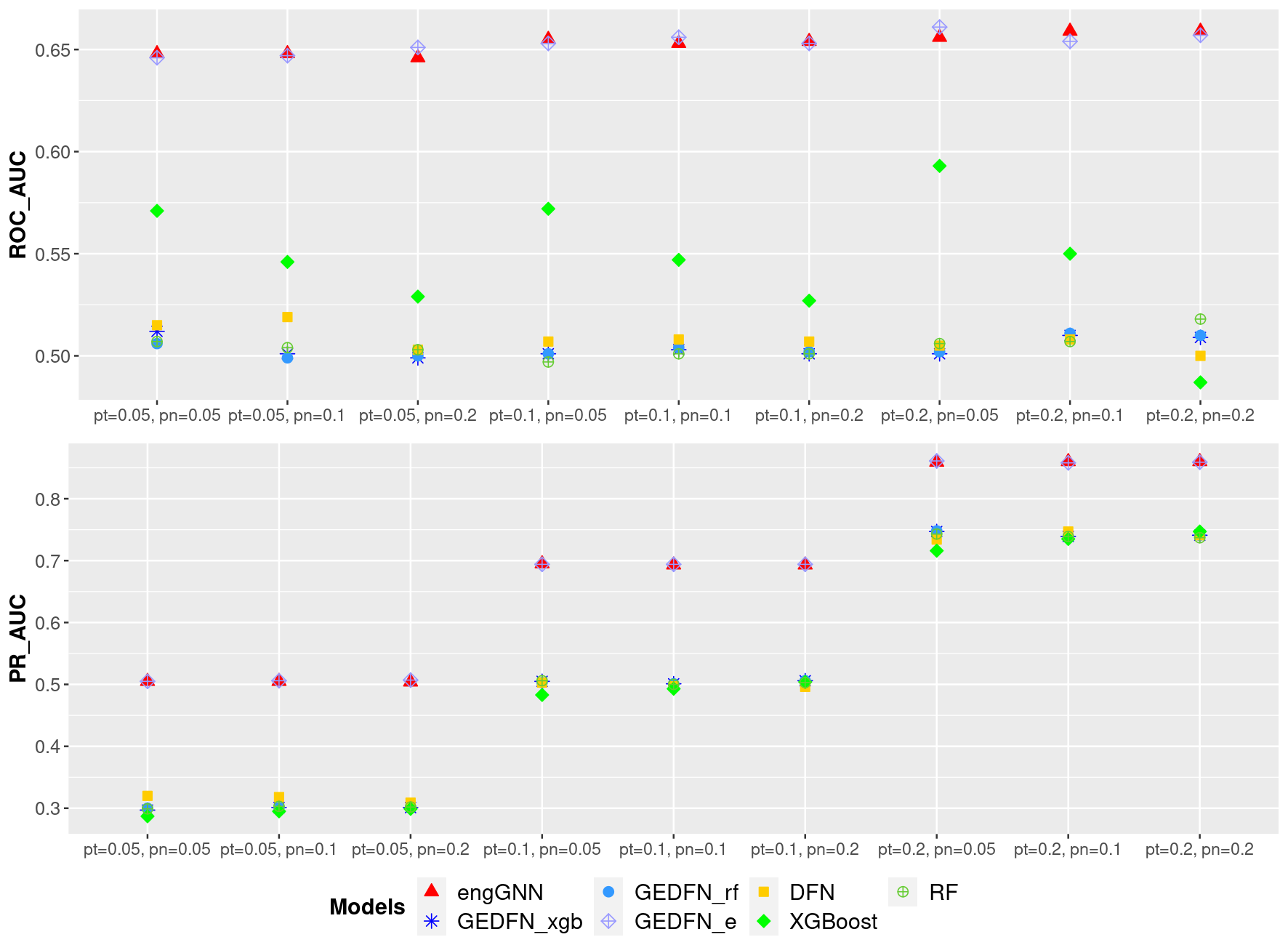}
\caption{Average feature selection performance of various models on simulated datasets across nine scenarios. \label{CombinedROCAUC_PRAUC_9scenarios}}
\end{figure*}
\clearpage

\begin{figure*}[h]
\centering
    \addtolength{\leftskip} {-2.6cm} 
    \addtolength{\rightskip}{-2.6cm} 
\includegraphics[width=0.8\textwidth]{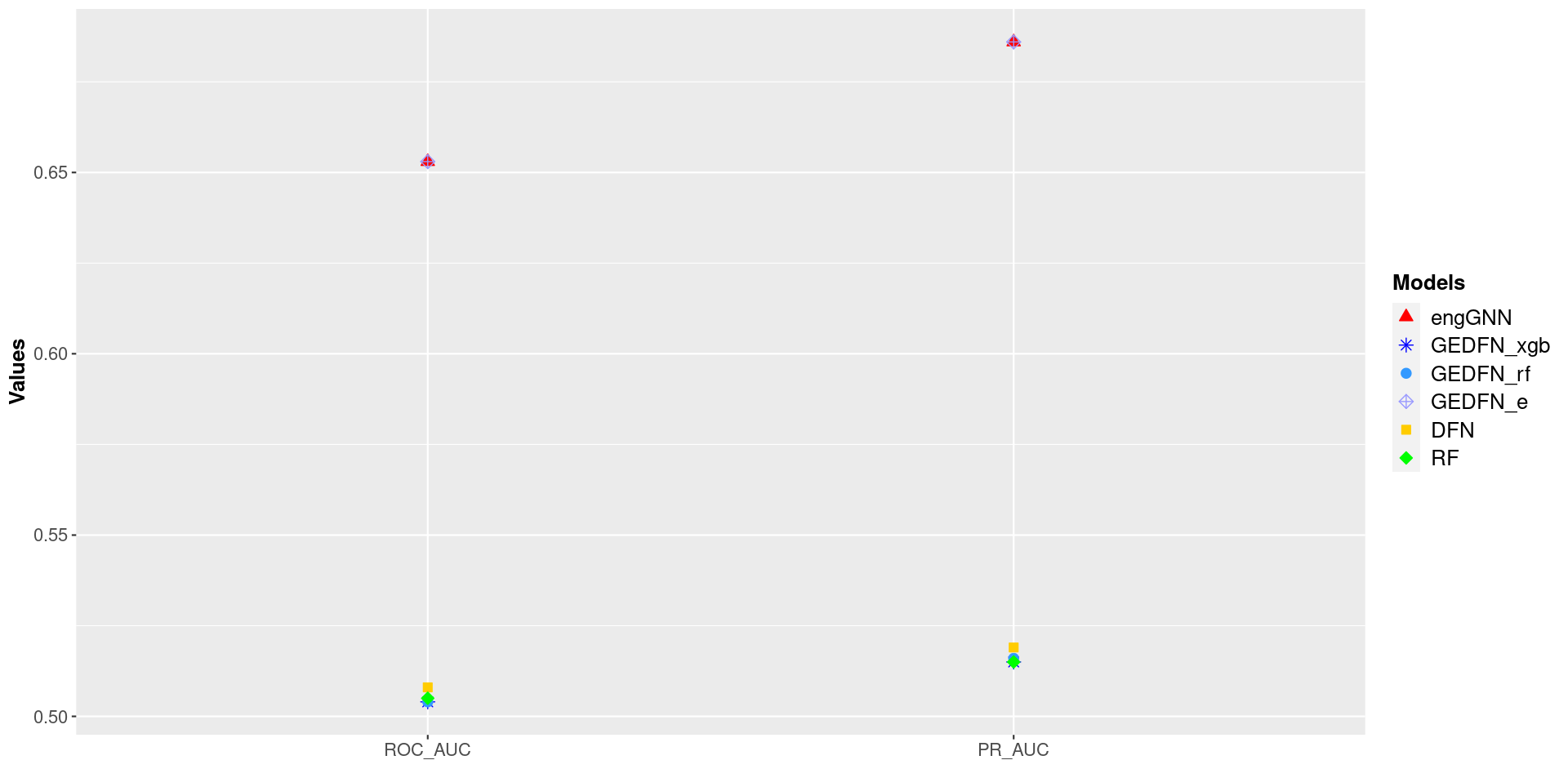}
\caption{Average feature selection ROC-AUC and PR-AUC of various models on simulated datasets across all the scenarios.  \label{all_roc-auc_pr-auc_FeatureSelection}}
\end{figure*}
\clearpage


\begin{figure*}[h]
\centering
    \addtolength{\leftskip} {-2.6cm} 
    \addtolength{\rightskip}{-2.6cm} 
\includegraphics[width=0.9\textwidth]{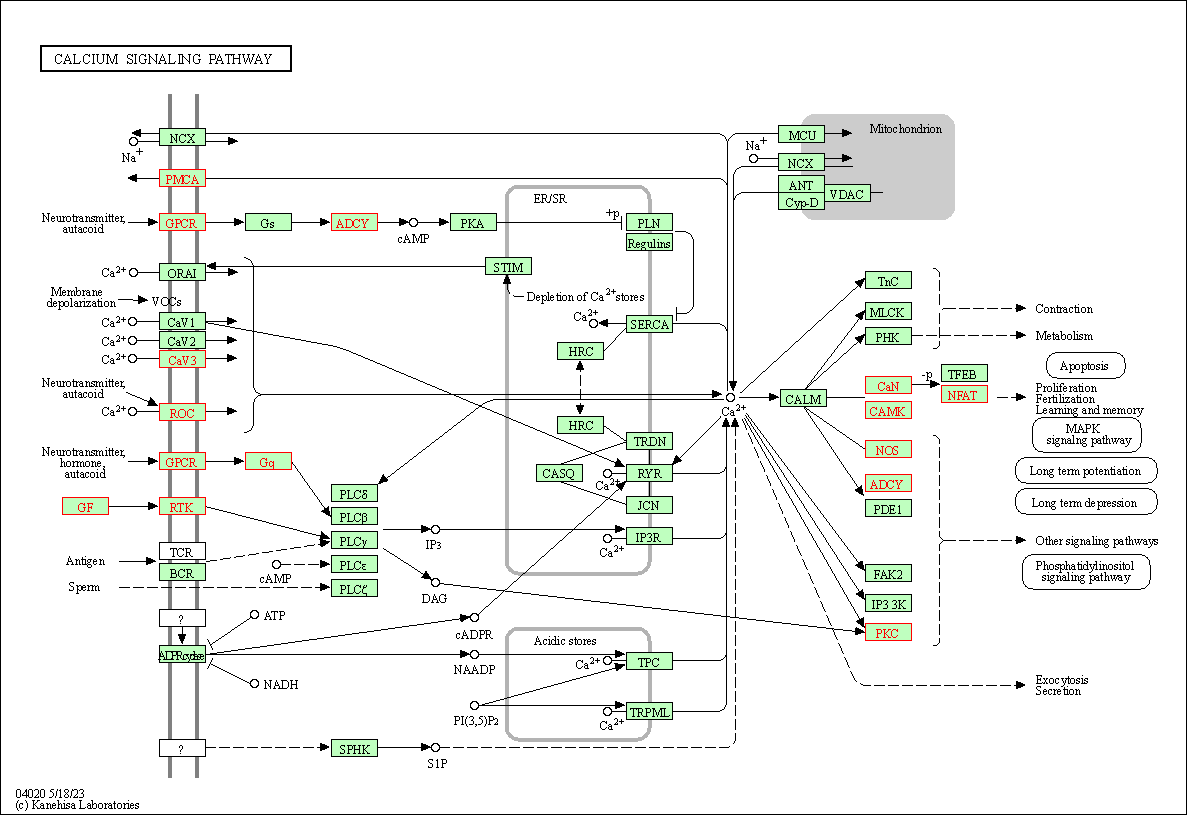}
\caption{Calcium signaling pathway \label{hsa04020}}
\end{figure*}
\clearpage

\begin{figure*}[h]
\centering
    \addtolength{\leftskip} {-2.6cm} 
    \addtolength{\rightskip}{-2.6cm} 
\includegraphics[width=0.9\textwidth]{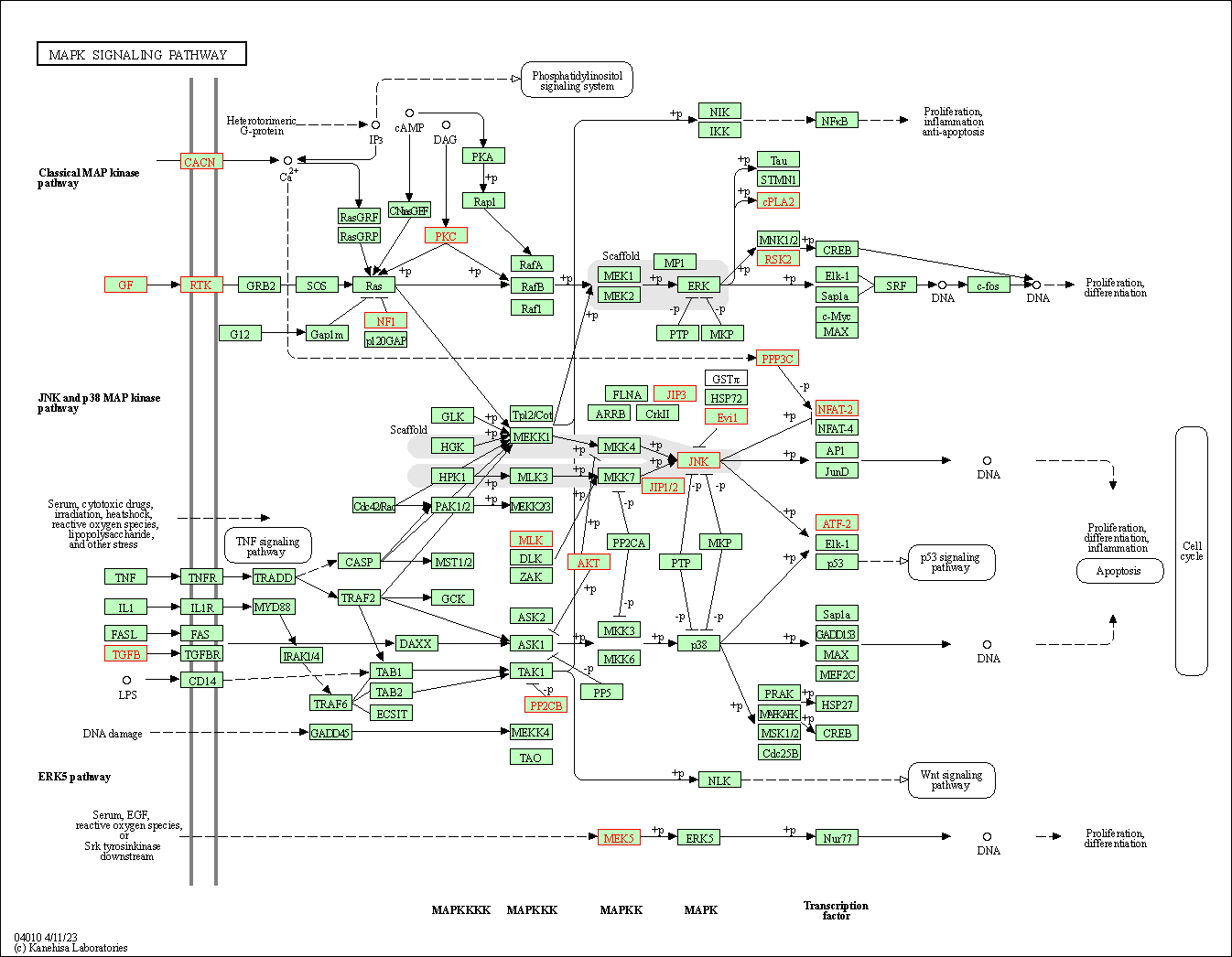}
\caption{MAPK signaling pathway \label{hsa04010}}
\end{figure*}
\clearpage

\begin{figure*}[h]
\centering
    \addtolength{\leftskip} {-2.6cm} 
    \addtolength{\rightskip}{-2.6cm} 
\includegraphics[width=0.9\textwidth]{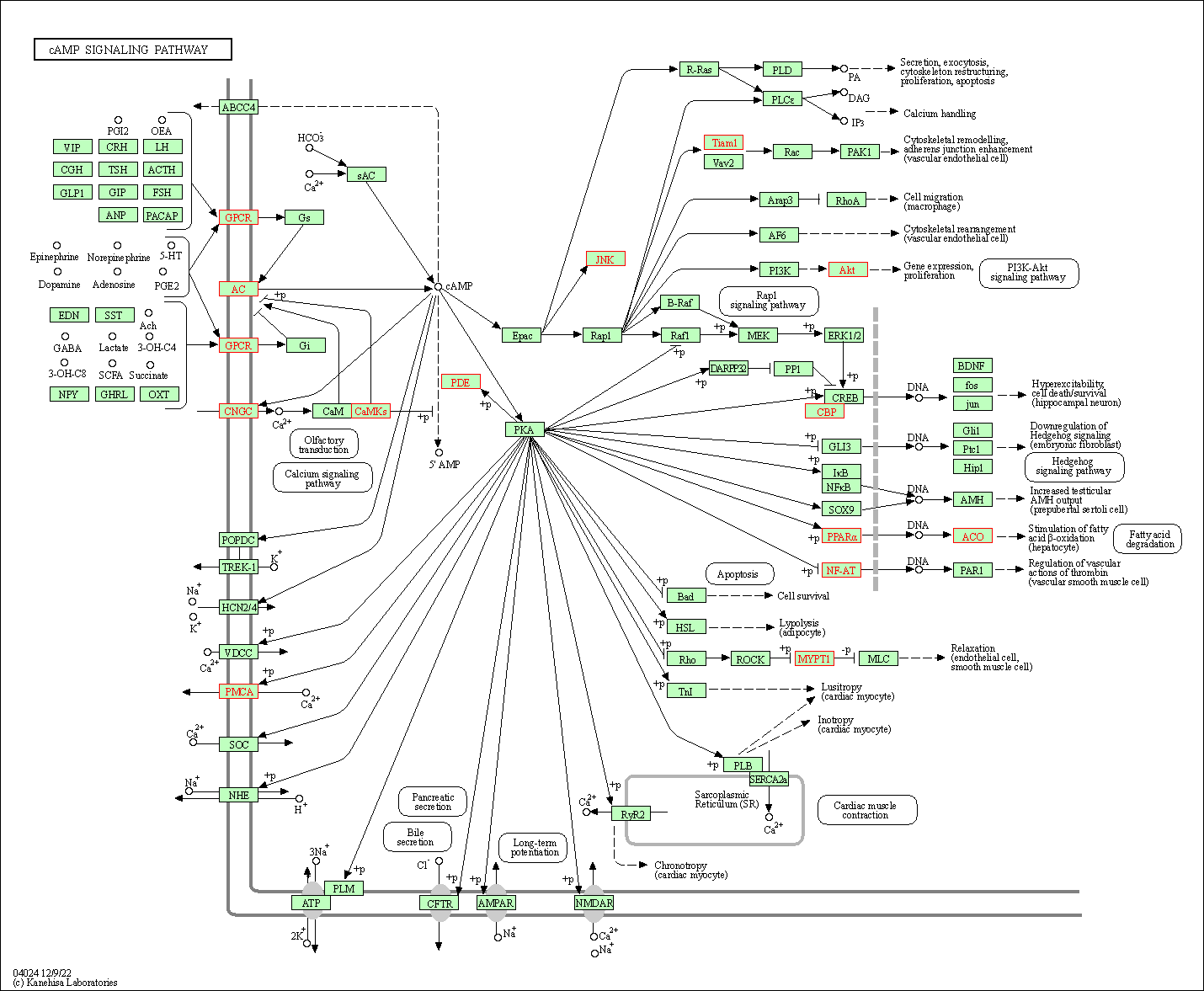}
\caption{cAMP signaling pathway \label{hsa04024}}
\end{figure*}
\clearpage


\begin{table*}[h]
\caption{Hyperparameter tuning for engGNN on Simulated and Real Data. The optimal choices are bolded. }
\label{tab:hyperparams}
\centering
\begin{tabular}{ll}
\toprule
Hyperparameter & Tuning Range \\
\midrule
Number of Trees in XGBoost & 100, 1000, $\mathbf{0.2p}$ \\
Hidden neural size in GNN & ($p$, 64), ($p$, 128), (\textbf{$p$, 64, 16}), ($p$, 64, 32) \\
Hidden neural size in DFN & 8, \textbf{16}, 32 \\
Activation Function & \textbf{ReLU} \\
Learning Rate & \textbf{0.0001} \\
Batch Size & 8, \textbf{16}, 32 \\
Training Epochs & \textbf{50}, 100 \\
Dropout Rate & \textbf{0.2}, 0.5 \\
\bottomrule
\end{tabular}
\end{table*}

\begin{table*}[h]
\caption{Feature selection performance (ROC-AUC and PR-AUC) across simulation scenarios with varying $p_t$ (true feature proportion) and $p_n$ (sample-to-feature ratio). The best-performing values in each row are bolded.}
\label{tab:feature_selection_combined}
\centering
\begin{adjustbox}{max width=\textwidth}
\begin{tabular}{c|cc|ccccccc}
\toprule
Metric & $p_t$ & $p_n$ & engGNN & $\mathrm{GEDFN}_{\text{xgb}}$ & $\mathrm{GEDFN}_{\text{rf}}$ & $\mathrm{GEDFN}_{\text{e}}$ & DFN & XGBoost & RF \\
\midrule
\multirow{9}{*}{ROC-AUC} 
 & 0.05 & 0.05 & \textbf{0.648} & 0.512 & 0.506 & 0.646 & 0.515 & 0.571 & 0.507 \\
 & 0.05 & 0.10 & \textbf{0.648} & 0.501 & 0.499 & 0.647 & 0.519 & 0.546 & 0.504 \\
 & 0.05 & 0.20 & 0.646 & 0.499 & 0.500 & \textbf{0.651} & 0.503 & 0.529 & 0.503 \\
 & 0.10 & 0.05 & \textbf{0.655} & 0.501 & 0.501 & 0.653 & 0.507 & 0.572 & 0.497 \\
 & 0.10 & 0.10 & 0.653 & 0.503 & 0.504 & \textbf{0.656} & 0.508 & 0.547 & 0.501 \\
 & 0.10 & 0.20 & \textbf{0.654} & 0.501 & 0.502 & 0.653 & 0.507 & 0.527 & 0.501 \\
 & 0.20 & 0.05 & 0.656 & 0.501 & 0.502 & \textbf{0.661} & 0.505 & 0.593 & 0.506 \\
 & 0.20 & 0.10 & \textbf{0.659} & 0.510 & 0.511 & 0.654 & 0.508 & 0.550 & 0.507 \\
 & 0.20 & 0.20 & \textbf{0.659} & 0.509 & 0.510 & 0.657 & 0.500 & 0.487 & 0.518 \\
\midrule
\multirow{9}{*}{PR-AUC}   
 & 0.05 & 0.05 & \textbf{0.505} & 0.297 & 0.300 & \textbf{0.505} & 0.320 & 0.287 & 0.299 \\
 & 0.05 & 0.10 & 0.505 & 0.301 & 0.301 & \textbf{0.506} & 0.318 & 0.295 & 0.303 \\
 & 0.05 & 0.20 & 0.504 & 0.301 & 0.302 & \textbf{0.507} & 0.309 & 0.299 & 0.302 \\
 & 0.10 & 0.05 & \textbf{0.695} & 0.505 & 0.503 & 0.694 & 0.503 & 0.483 & 0.506 \\
 & 0.10 & 0.10 & 0.693 & 0.501 & 0.501 & \textbf{0.694} & 0.497 & 0.493 & 0.500 \\
 & 0.10 & 0.20 & 0.693 & 0.506 & 0.506 & \textbf{0.694} & 0.496 & 0.504 & 0.503 \\
 & 0.20 & 0.05 & 0.859 & 0.747 & 0.748 & \textbf{0.861} & 0.734 & 0.716 & 0.743 \\
 & 0.20 & 0.10 & \textbf{0.860} & 0.739 & 0.739 & 0.858 & 0.747 & 0.735 & 0.739 \\
 & 0.20 & 0.20 & \textbf{0.860} & 0.741 & 0.741 & 0.859 & 0.741 & 0.747 & 0.737 \\
\bottomrule
\end{tabular}
\end{adjustbox}
\end{table*}

\end{document}